# ELIXR: Towards a general purpose X-ray artificial intelligence system through alignment of large language models and radiology vision encoders

Shawn Xu[1,*], Lin Yang[1,*], Christopher Kelly[1,*], Marcin Sieniek[1], Timo Kohlberger[1], Martin Ma[1], Wei-Hung Weng[1], Atilla P. Kiraly[1], Sahar Kazemzadeh[1], Zakkai Melamed[1], Jungyeon Park[1], Patricia Strachan[1], Yun Liu[1], Chuck Lau[2], Preeti Singh[1], Christina Chen[1], Mozziyar Etemadi[3], Sreenivasa Raju Kalidindi[4], Yossi Matias[1], Katherine Chou[1], Greg S. Corrado[1], Shravya Shetty[1], Daniel Tse[1], Shruthi Prabhakara[1], Daniel Golden[1], Rory Pilgrim[1], Krish Eswaran[1,‡], Andrew Sellergren[1,‡]

[1] Google LLC, Mountain View, CA, USA
[2] Work done at Google via Advanced Clinical, Deerfield, IL, USA
[3] Northwestern Medicine, Chicago, IL, USA
[4] Apollo Radiology International, Hyderabad, India
Address correspondence to: Andrew Sellergren (asellerg@google.com), Google LLC, 1600 Amphitheatre Parkway, Mountain View, CA 94043, USA

[*] Equal contributions.
[‡] Equal leadership.

# Abstract

**Background:** Artificial intelligence systems for medical imaging have traditionally focused on highly specific tasks and have generalized inconsistently to new problems. The combination of large language models (LLMs) and vision encoders offers the potential to address some of these challenges. In this work, we present an approach that enables efficient training of multimodal models using routinely collected medical images and their associated text reports, and adds the ability to perform a diverse range of tasks with rich expressive outputs. This approach unlocks the potential for a new generation of medical AI applications, supporting workflows including high performance zero-shot and data-efficient classification, semantic search, visual question answering (VQA), and radiology report quality assurance (QA).

**Methods:** Our approach, which we call Embeddings for Language/Image-aligned X-Rays, or "ELIXR", leverages a language-aligned image encoder "grafted" via an adapter onto a fixed LLM, PaLM 2, to perform a broad range of tasks. We train this lightweight adapter architecture using images paired with corresponding free-text radiology reports from the MIMIC-CXR dataset. Evaluation of zero-shot and data-efficient classification was performed using the public CheXpert and ChestX-ray14 datasets, as well as a private dataset from five hospitals in India. Semantic search was evaluated across four themes using the MIMIC-CXR test set. VQA was evaluated using the VQA-RAD benchmark and the MIMIC-CXR test set. LLM output for report QA was evaluated on the MIMIC-CXR test set by a board-certified thoracic radiologist.

**Results:** ELIXR achieved state-of-the-art performance on zero-shot chest X-ray (CXR) classification (mean AUC of 0.850 across 13 findings), data-efficient CXR classification (mean AUCs of 0.893 and 0.898 across five findings (atelectasis, cardiomegaly, consolidation, pleural effusion, and pulmonary edema) for 1% (~2,200 images) and 10% (~22,000 images) training data), and semantic search (0.76 normalized discounted cumulative gain (NDCG) across nineteen queries, including perfect retrieval on twelve of them). Compared to existing data-efficient methods including supervised contrastive learning (SupCon), ELIXR required two orders of magnitude less data to reach similar performance. ELIXR also showed promise on CXR vision-language tasks, demonstrating overall accuracies of 58.7% and 62.5% on visual question answering and report quality assurance tasks, respectively. These results suggest that ELIXR is a robust and versatile approach to CXR AI.

**Conclusion:** LLM-aligned multimodal models can unlock the value of chest X-rays paired with radiology reports to solve a variety of previously challenging tasks.

# Introduction

The past decade has witnessed dramatic advances in artificial intelligence (AI) in medical imaging. Numerous deep learning systems have been developed that can achieve expert-level performance across a range of medical tasks[1]. However, clinical and technical limitations have resulted in an implementation gap that has impeded the impact of AI in real world health applications at scale[2]. Key challenges include the significant cost of curating high quality training datasets, restriction of AI development to narrow, highly specific tasks, difficulty in processing multimodal data, and limited interpretability that has hampered effective human-AI interaction[3].

Until recently, AI systems for medical imaging have been largely built using vision-only models, including convolutional neural networks (CNNs) and vision transformers[4]. Using a traditional fully supervised approach, training CNNs and vision transformers is an expensive and time-consuming process that requires large quantities of expertly annotated data[5]. In addition, such networks are usually limited to performing discrete tasks, such as image classification, object detection, and segmentation. On the input side, these networks also take in only images, usually of just one modality. In contrast, healthcare workflows are typically multimodal in nature, with clinicians leveraging a diverse array of inputs (e.g. clinical notes, images, investigations) when making diagnoses and treatment decisions.

Large language models (LLMs) are part of a new generation of versatile transformer-based[6] AI models that are trained on massive datasets and demonstrate previously unseen abilities to generalize to a range of tasks, despite requiring very little task-specific data[7,8]. The multimodal combination of vision models and LLMs presents a range of exciting possibilities including zero-shot image-to-text generation that can follow natural language instructions[9–11]. In medical imaging, these advances offer the potential to address limitations of vision-only models by enabling model training using ubiquitous medical images that have paired free text reports, adding capabilities to carry out a diverse range of tasks, accurately coping with the long-tail of diagnoses, enabling true multimodal inference, and presenting new options for expressive human-computer interaction[7].

In this paper, we present a lightweight vision-language adapter model called ELIXR (Embeddings for Language/Image-aligned X-Rays), which builds upon prior work[9–11] to combine or "graft" a vision encoder with a frozen LLM to perform a wide range of vision-language tasks broadly relevant to medical imaging. Our case study in this work focuses on chest X-rays (CXRs) due to the wide availability of image-text paired data, but the methods are applicable to other image modalities. We describe the following key advantages of ELIXR:

1. **ELIXR achieves state-of-the-art performance for zero-shot classification, data-efficient classification, and semantic search of thoracic conditions across a range of datasets.** These factors may enable a new class of models that can address the long-tails of diagnoses in the medical domain, and provide a path for more broadly useful AI tools in the diagnostic workflow.

2. **Adapting an image encoder to an LLM using ELIXR is a fast and resource-efficient method of training compared to full finetuning of an LLM**, leveraging a modest-sized frozen LLM and a data-efficient frozen vision encoder. Building models on top of ELIXR can be done rapidly to prototype new use cases, adapt to distribution shifts with a small amount of new training data, or use alternative publicly available LLMs.

3. **ELIXR's synthesis of imaging and text unlocks a new generation of medical AI applications.** In this study we demonstrate semantic search, visual question answering (VQA), and radiology report quality assurance (QA), but there are countless potential applications across the medical domain that can be addressed using the proposed multimodal framework.

4. **ELIXR is trained using paired CXR and free text radiology reports - data that are ubiquitous in healthcare.** The training process does not require expensive manual label curation by experts. Such an approach unlocks the value of routinely collected medical data to develop AI systems at far greater scale and at lower overall cost than previously possible.

# Methods

## ELIXR system

We trained the ELIXR system in two stages (*ELIXR-C* and *ELIXR-B*).

First, we trained the *ELIXR-C* model using Contrastive Language–Image Pre-training (CLIP)[9] (Figure 1a). This uses radiology reports to align our previously published pre-trained supervised contrastive learning-based (SupCon) vision-only CXR model[12] with a T5 language encoder[13]. CLIP uses a contrastive loss function, which encourages the model to bring the representations of an image and its associated text (in this case, the radiology report) closer together in a high-dimensional space, while simultaneously pushing apart representations of mismatched images and text.

Second, we trained an LLM-aligned adapter network *ELIXR-B* (Figure 1b), based on the Bootstrapping Language-Image Pre-training 2 (BLIP-2) architecture[10]. *ELIXR-B* is built directly upon *ELIXR-C*, where it aims to extract location-aware features from the unpooled spatial *ELIXR-C* image embedding space and map them to the LLM's language token space. In this work, we used PaLM 2-S as the LLM[14]. By serving as an adapter between the image encoder and the LLM, *ELIXR-B* passes information between vision and language encoders via an attention mechanism, and allows us to leverage the existing knowledge and reasoning abilities of the LLM to interpret the images and perform various vision-language tasks (e.g. captioning, VQA). For computation and data efficiency, we keep both *ELIXR-C* and PaLM 2-S frozen, and only train the adapter between them. This can be thought of as a way of grafting an image encoder onto an LLM. More specifically, following the BLIP-2 architecture[10], there are two phases to *ELIXR-B* training: vision-language representation learning (phase 1) and vision-language generative learning (phase 2). In the first phase, the vision-language model (the Q-Former) is trained to understand and represent both CXRs and reports in a shared embedding space by jointly employing three different tasks: a) image-text contrastive learning (ITC), b) image-grounded text generation (ITG), and c) image-text matching (ITM). Standard contrastive loss is applied for image-text contrastive learning; image-grounded text generation is modeled as a classification problem (i.e. which token in the vocabulary should be chosen at each output position) and optimized by cross-entropy loss; image-text matching is modeled as a binary classification problem (image-text matched/unmatched) and optimized by cross-entropy loss. This results in a model that can extract key information from the image embeddings and align it with the report text embedding space. In the second phase, a multilayer perceptron (MLP) that connects the Q-Former with the LLM, and the Q-Former itself are further trained to generate the impressions section of radiology reports based upon the image embeddings from ELIXR-B using the LLM. The language modeling (standard cross-entropy) loss is used to guide the training. The result is that the Q-Former is able to produce LLM-aligned tokens based on the image and feed the most useful information to the LLM, while removing irrelevant visual information.

(a)

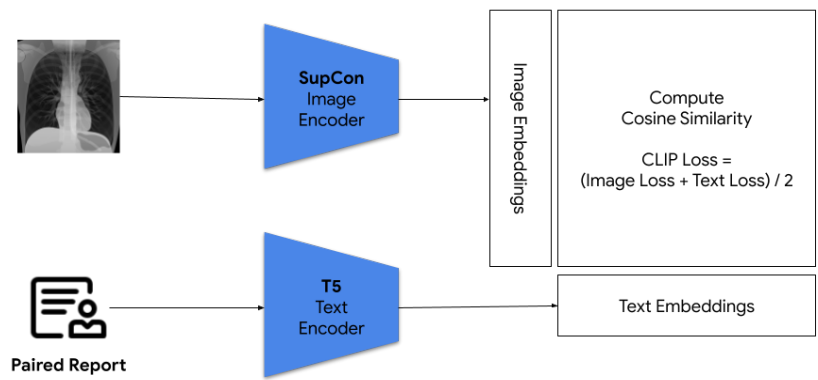

**ELIXR-C Training**

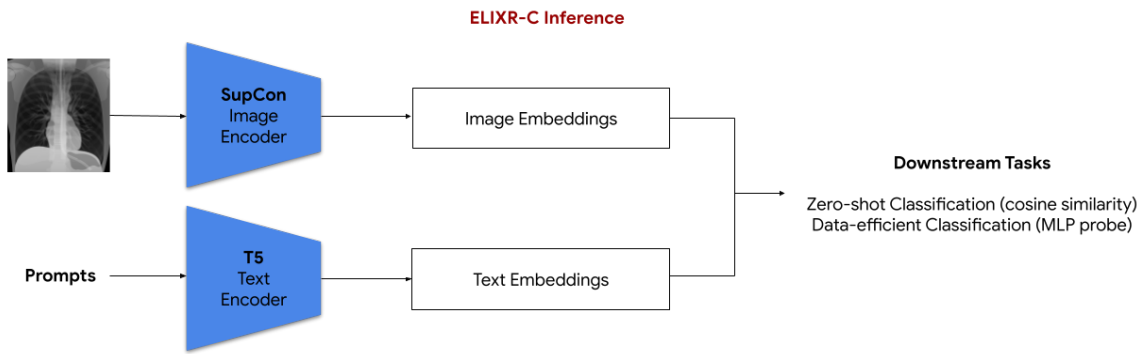

**ELIXR-C Inference**

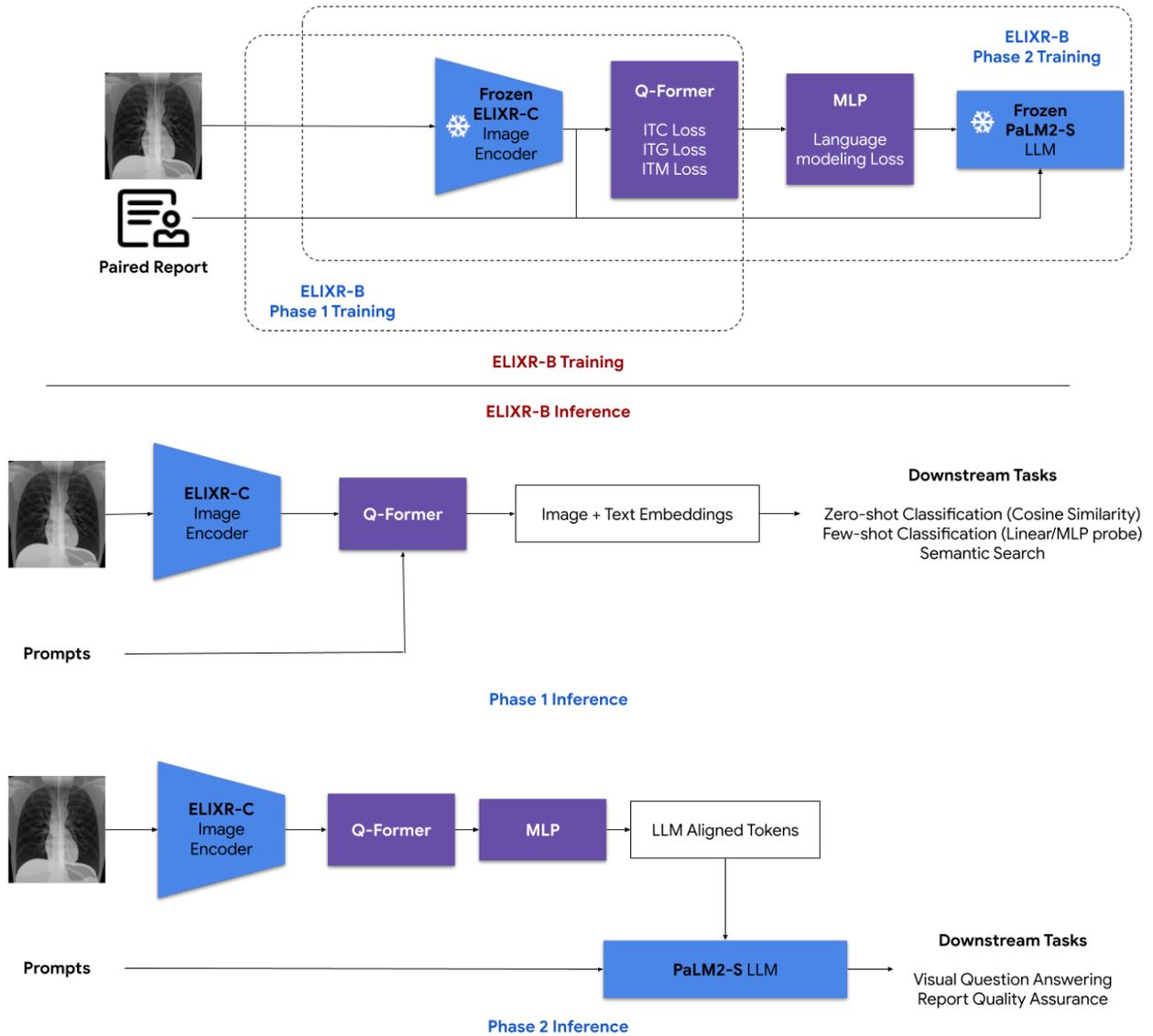

**Figure 1: Architecture of *ELIXR*.** (a) Training and inference of *ELIXR-C*. (b) Training and inference of *ELIXR-B*. *ELIXR-B* is trained in two phases. In the first phase, the model bootstraps vision-language representation in the Q-Former with three learning objectives (image-text contrastive learning (ITC), image-grounded text generation (ITG), image-text matching (ITM) losses) to learn from embeddings from a frozen image encoder. In the second phase, the model bootstraps vision-to-language generation from a frozen large language model. The purple text boxes represent the learned (unfrozen) components in the training step. Details of the VQA inference are further described in the relevant section.

## Datasets

We included more than one million CXR images from six datasets in this study as described in Table 1: five hospitals from India (IND1), a hospital from Illinois, USA (US) (US1), Beth Israel Deaconess Medical Center in Massachusetts, USA (MIMIC-CXR), National Institutes of Health (NIH) Clinical Center in Maryland, USA (CXR-14), Stanford Health Care in California, USA

(CheXpert), and the VQA-RAD dataset from the National Library of Medicine MedPix platform[15–21].

Data from IND1, US1, and the MIMIC-CXR train set were used for training *ELIXR-C*, while only data from the MIMIC-CXR train set were used for training *ELIXR-B*. Details of datasets and the evaluation tasks in which they were used appear in Table 1. Labels for IND1 and US1 are as previously described[12].

| Dataset | IND1 | US1 | MIMIC-CXR | CXR-14 | CheXpert | VQA-RAD (Chest-only) |
|---|---|---|---|---|---|---|
| **Dataset usage** | | | | | | |
| Development (train/validation sets) | *ELIXR-C* | *ELIXR-C* | *ELIXR-C* and *ELIXR-B* | | | |
| Evaluation (test set) | Data-efficient and zero-shot | | All tasks | Data-efficient and zero-shot | Data-efficient and zero-shot | VQA |
| **Dataset and patient statistics** | | | | | | |
| Dataset origin | Five hospitals in India | An AMC in Illinois, USA | AMC in Massachusetts, USA | NIH, Maryland, USA | AMC in California, USA | NIH, Maryland, USA |
| Number of patients | 348,335 | 12,988 | 60,523 | 30,805 | 65,654 | 107 |
| Age (IQR) | 35-58 | 48-71 | 43-72 | 34-59 | N/A | N/A |
| Sex | Female: 133,833 (38.5%) Male: 214,334 (61.5%) Unknown: 168 (< 0.1%) | Female: 6,779 (52.2%) Male: 6,209 (48.8%) | Female: 31,610 (52.2%) Male: 28,913 (48.8%) | Female: 14,175 (46.0%) Male: 16,630 (54.0%) | N/A | N/A |
| **Image and finding statistics** | | | | | | |
| Number of images | 485,082 | 165,182 | 243,324 | 104,278 | 223,648 | 107 |
| View (AP/PA) | AP: 79,958 (16.5%) PA: 625,735 (83.5%) | AP: 108,822 (65.9%) PA: 24,269 (14.7%) Unknown: 32,091 (19.4%) | AP: 147,169 (60.4%) PA: 96,155 (39.6%) | AP: 44,811 (40.0%) PA: 67,305 (60.0%) | N/A | N/A |
| Airspace opacity | 43,629 (9.0%) | 15,309 (10.1%)* | 54,769 (22.5%) | 3,485 (3.3%) | 94,328 (30.2%) | N/A |
| Fracture | 5,200 (1.1%) | 5,760 (3.8%)* | 4,781 (2.0%) | 546 (0.5%) | 7,436 (2.4%) | N/A |
| Pneumothorax | 1,657 (0.3%) | 7,202 (4.8%)* | 11,235 (4.6%) | 5,302 (5.1%) | 17,700 (5.7%) | N/A |
| Consolidation | 15,144 (3.1%) | 6,315 (4.2%)* | 11,525 (4.7%) | 4,667 (4.5%) | 13,015 (4.2%) | N/A |
| Pleural effusion | 1,228 (0.3%) | 33,280 (22.0%*) | 57,721 (23.7%) | 13,317 (12.8%) | 76,963 (24.6%) | N/A |
| Pulmonary edema | 1,136 (0.2%) | 34,301 (22.7%)* | 29,331 (12.1%) | 2,303 (2.2%) | 49,717 (15.9%) | N/A |
| Atelectasis | 15,929 (3.3%) | 49,293 (32.6%)* | 48,790 (20.1%) | 11,559 (11.1%) | 29,795 (9.5%) | N/A |
| Cardiomegaly | 1,115 (0.2%) | 17,001 (11.3%)* | 47,673 (19.6%) | 2,776 (2.7%) | 23,451 (7.5%) | N/A |
| Support Devices | 29,698 (6.1%)* | 97,463 (64.6%)* | 73,294 (30.1%) | N/A | 107,014 (56.2%) | N/A |
| Enlarged cardiomediastinum | 349 (0.1%)* | 421 (0.3%)* | 7,657 (3.1%) | N/A | 9,273 (4.9%) | N/A |
| Lung lesion | 7,713 (1.6%)* | 1351 (0.9%)* | 6,632 (2.7%) | N/A | 7,022 (3.7%) | N/A |
| Pleural other | 19,301 (4.0%)* | 1807 (1.2%)* | 2,083 (0.9%) | N/A | 2,493 (1.3%) | N/A |
| Pneumonia | 54 (0.0%)* | 29,816 (19.7%)* | 17,222 (7.1%) | 1,255 (1.2%) | 4,657 (2.4%) | N/A |

**Table 1: Descriptive statistics of the datasets used in the study.** *: estimated from radiology reports

# Evaluation

We demonstrated the utility of ELIXR on five CXR-related tasks: (1) zero-shot classification, (2) data-efficient classification, (3) semantic search, (4) VQA, and (5) report QA (Table 2). Zero-shot and data-efficient image classification as well as semantic search were performed using *ELIXR-C* and *ELIXR-B* phase 1 (language-aligned image embeddings), while VQA and quality assurance were performed using *ELIXR-B* phase 2, which combined these embeddings with the fixed PaLM 2-S LLM[14].

| Task | Input | Model output | Metric | ELIXR versions used |
|---|---|---|---|---|
| **Zero-shot classification** | CXR image<br>Positive prompt(s)<br>Negative prompt(s) | Classification score | AUC | *ELIXR-C* and *ELIXR-B* phase 1 |
| **Data-efficient classification** | For training small nonlinear classifiers on embeddings: variable amount of CXR images with their corresponding annotations<br>For inference: CXR image | Classification score | AUC | *ELIXR-C* and *ELIXR-B* phase 1 |
| **Semantic search** | Text description of search term | Top-5 CXR images from MIMIC-CXR test set that are related to the description | NDCG@5 Precision | *ELIXR-C* and *ELIXR-B* phase 1 |
| **Visual question answering (VQA)** | CXR image<br>Questions about the image | Answers to given questions based on the image | Accuracy | *ELIXR-B* phase 2 |
| **Report quality assurance (QA)** | CXR image<br>Normal/altered radiology report | Decision about accuracy of report, along with rich text explanation. | Accuracy | *ELIXR-B* phase 2 |

**Table 2: Downstream chest X-ray tasks that are evaluated in this study**. *ELIXR-C* takes images and/or text as input and outputs embeddings; *ELIXR-B* takes images and/or text as input and outputs embeddings and/or text. Therefore, for text output tasks such as VQA and QA, only *ELIXR-B* is used.

## Classification

For zero-shot and data-efficient classification of CXR findings, area under the receiver operating characteristic curve (AUC) for each finding class is reported since the tasks are all binary classifications. We compared *ELIXR-C* and *ELIXR-B* performance with SupCon and previous SOTA models CheXzero and ConVIRT as baselines[22,23]. We also evaluated the effect of varying the training dataset size.

**Zero-shot classification**

To perform zero-shot classification using *ELIXR-C*, we adopted a prompting strategy described previously[23]. Note that because *ELIXR-C* was pretrained on images and reports that include the findings being classified, "zero-shot" refers more to open-ended classification without using explicit labels during training than it does to never having observed these findings during training. The positive and negative text prompts were passed through the text encoder to obtain their text embeddings. These text embeddings were each average pooled and normalized, resulting in one representative embedding for each prompt. A cosine similarity was then calculated between the image embedding and these two text embeddings. A softmax of the cosine similarity for the positive prompt was used to produce a classification score so that AUC can be calculated. When there are multiple prompts for the positive or negative case, the mean of the cosine similarities of these prompts was taken before computing the softmax. Details of the prompts are listed in Supplementary table 1.

For *ELIXR-B*, the image as well as positive and negative text prompts were passed through the Q-Former to get the output image query tokens and the sequence of text embeddings. The first in the sequence of text embeddings is the single special classification token (`[CLS]`). The cosine similarity is calculated between this classification token and each of the output image query tokens. The highest cosine similarity for the positive prompt and the highest cosine similarity for the negative prompt are passed to the softmax function to get the final classification score.

**Data-efficient classification**

To perform data-efficient classification, we followed the same procedure as in Sellergren et al[12]: a nonlinear classifier, an MLP consisting of two layers of 512 and 256 neurons, was trained on top of the frozen image encoder. We adopted a learning rate of 0.2, batch size of 512, and trained for 300 epochs with the layer-wise adaptive rate scaling (LARS) optimizer. To make results more directly comparable to CheXzero and ConVIRT, we also trained linear classifiers for the 1% and 10% training data samples.

**Statistical analysis**

For data-efficient classification, the AUC was averaged across 10 repeats of randomly subsampled training sets. To obtain an overall summary, the AUCs were (also) averaged across all tasks and all datasets (CheXpert, CXR-14, and IND1) for each training set size. 95% confidence intervals were calculated using twice the standard error across the 10 randomly drawn train set samples, and hypothesis testing was based on comparing the model closest to the mean performance across the 10. For the zero-shot models, 95% confidence intervals were estimated and AUCs were compared using the DeLong method[24].

For semantic search, confidence intervals were calculated by bootstrapping with 1,000 samples. p-values were calculated from two-sided permutation tests with 1,000 iterations.

Semantic search

For semantic search (also known as text-image retrieval), we provided queries across four topics, including queries for single findings, laterality-specific, severity-specific, and nuanced features. For each query, the analysis focused on the top five retrieved images based on model predictions. There were seven single finding queries, three laterality queries, four severity queries, and five nuanced queries. We compared *ELIXR-C* and *ELIXR-B* against the current state-of-the-art model MedCLIP[25] using the publicly available code and model checkpoints. The MIMIC-CXR test set served as the data pool for retrieval. The full list of 19 queries is provided in Supplementary table 2.

For semantic search using *ELIXR-C*, we computed the cosine similarity between the query text embedding and the image embeddings from the data pool. The top five images with the highest cosine similarity were retrieved. For semantic search using *ELIXR-B*, we adopted the two-stage method in BLIP-2[10]. In the first stage, we retrieved 128 images with the highest cosine similarity similar to CLIP as the candidates. In the second stage, the image-text matching score (i.e. the matched class probability) was computed to rerank these 128 candidates. The top five images with the highest image-text matching score from these 128 candidates were then returned.

For evaluation, a board-certified thoracic radiologist (CL) scored the semantic search results as follows: 0 = irrelevant or factually incorrect, 1 = close fit to the query, but not the one intended (e.g., retrieved "bilateral effusion" for "right effusion" query), and 2 = reasonable match. Precision at five (precision@5) and normalized discounted cumulative gain at 5 (NDCG@5) were calculated to evaluate the quality of retrieval[26]. For the ideal rank normalization, we assumed the data pool contained at least five matched queries, and thus we set the relevance scores to be all 2s for the ideal relevance. Precision for cases that had reasonable match (score = 2) and precision for at least somewhat fit (score ≥1) were also measured.

To establish the context of the difficulties of the retrieval, we estimated the total count of images with corresponding pathologies. To do so, we adopted an LLM-based approach to detect mislabeling within the MIMIC-CXR test set[27,28]. Candidates for correction were first identified by a keyword search on the radiology reports. Next, a medically tuned LLM (Med-PaLM 2[29]) was applied to ensure that the label was consistent with the report, and a board-certified thoracic radiologist (CL) adjudicated cases where the LLM results differed from the ground truth in MIMIC-CXR. Details are listed in Supplementary table 3 and Supplementary figure 1.

## Visual question answering

We used two different datasets for VQA evaluation of *ELIXR-B*: a subset of the MIMIC-CXR test set, which is from the same domain as a part of the training dataset, and the subset of chest x-ray cases (`IMAGEORGAN == "CHEST" AND ANSWER != "ultrasound"`) in the VQA-RAD dataset[17], from a different domain than the training dataset. A small held-out tuning set consisting only of "yes" or "no" answers from VQA-RAD was used for model checkpoint selection. Specifically, we used our own test and tuning splits for the 793 VQA-RAD CXR question and answer pairs, since the original test set (column `Phrase_type/QID_para == "test_freeform"` or `"test_para"`) shares images with the development set (`QID_para == "freeform"` or `"para"`). By contrast, the images in our test set (584 questions on 73 images) are disjoint from the tune set images, and thus were unseen by the model until evaluation time. Moreover, we did not train the model on any VQA-RAD question and answer pairs, but only used it for checkpoint selection based on our smaller tuning set of 209 VQA pairs across 25 images. We compared *ELIXR-B* against the SOTA MedVInT[30] model using the publicly available code and a model checkpoint that, like ours, was not finetuned on VQA-RAD.

The second VQA dataset is a subset of cases with findings from MIMIC-CXR test set, as labeled in the MIMIC-CXR-JPG project[27,28], with eight cases for each for the following findings: "No Finding", "Pneumothorax", "Pleural Effusion", "Edema", "Consolidation OR Pneumonia" (collected as a single category), and "Lung Lesion". See the report quality assurance section below for further details on the case selection. For each case with a finding present (with finding presence confirmed by a board-certified thoracic radiologist (CL)), we queried the model with a set of finding-specific questions, covering its presence, location, and its severity, size or type, where applicable. See Supplementary table 4 for the complete question catalog. In addition, we asked two additional finding-independent questions per case, and three finding-independent questions for cases without any of the above findings (category "finding-independent" in the Supplementary table 4).

Since both phases of *ELIXR-B* were trained to generate the impression section of a radiology report, but neither was tuned to follow instructions, we utilized both the impression generation results from phase 1 (same ITG setup as BLIP-2,[10]) and LLM-aligned tokens from phase 2 to facilitate the VQA use case. Specifically, after generating the impression with phase 1, we then ran inference for the phase 2 model, using the following dialog prompt for the PaLM 2-S LLM wherein the LLM-aligned tokens from phase 2 inference were placed at the beginning of the prompt at (`{aligned LLM tokens}`), the phase 1-generated impression was fed to (`{impression}`), and the specific question (`{question}`) was added afterwards:

```
{aligned LLM tokens}
[Bot] I'm a helpful Chest X-ray assistant, I can help you interpret the above image.
[User] What are the findings?
[Bot] {impression}
[User] Based on the above chest x-ray, the findings, and/or your
```

```
medical knowledge, answer the following question: {question}
[Bot]
```

In terms of answer grading, model answers that could programmatically be mapped to "yes" or "no" and compared against "yes"-or-"no"-expected answers were automatically graded with 1.0 for a match, and 0.0 otherwise. A board-certified thoracic radiologist (CL) graded the remaining model answers using correctness scores 0.0, 0.5 or 1.0. See Supplementary table 5 for the full scoring rubric.

For the MIMIC-CXR test set samples, the same radiologist using the same scoring rubric evaluated all 216 answers generated by the model based on their assessment directly of the CXR image and original radiology report.

Sensitivity, specificity and accuracy values were calculated from the average radiologist grades on respective subsets where the condition was positive, negative or both.

### Report quality assurance

In the report quality assurance task, we simulated the situation where a radiology report contained errors and used ELIXR-B to identify and suggest corrections to these errors. Errors that we evaluated included swapping laterality of a finding ("left" to "right" or vice versa), adding an erroneous finding that was not clearly present in the image, or omitting a major finding that was present in the image.

To evaluate ELIXR's performance, we first identified a subset of cases with findings from the MIMIC-CXR test set, as labeled in the MIMIC-CXR-JPG project, that would be most relevant for evaluating quality assurance: "No Finding", "Pneumothorax", "Pleural Effusion", "Edema", "Consolidation OR Pneumonia" (collected as a single category), and "Lung Lesion". We randomly selected eight cases per finding. For each case we defined the "primary finding" as the finding that was used in the query that yielded that given case, even if other findings, including more clinically significant findings, were present in the case. For example, if we searched for cases with pleural effusion and discovered a case with both pleural effusion and pulmonary edema present, the primary finding for that case was considered to be pleural effusion. We filtered cases to ensure that (1) an impression section was present in the report (true for approximately 83% of cases) and (2) the primary finding was present unambiguously in the image, based on the impression of a board-certified thoracic radiologist (CL).

Within the set of eight cases for each finding category, we made an alteration per case as follows: two cases we left with unaltered impression sections (the "control" cases), two cases had the primary finding's laterality swapped, two cases had the primary finding removed, and two cases had an erroneous finding added. For each alteration, we made further minimal modifications to the impression text as needed to ensure that it was internally consistent (e.g., if we added an extraneous finding of "moderate pulmonary edema," we confirmed that the rest of the report was clinically consistent with this alteration). A summary of these alterations appears in Supplementary table 6. For "Edema", because pulmonary edema tends to be a bilateral finding, we did not include any cases with swapped laterality and instead included three control cases, three cases where an extraneous finding was added, and two cases where the primary finding of "Edema" was removed, still resulting in eight cases total. For the cases labeled as "No Finding", modifications like laterality swapping and finding removal were not feasible. Therefore, we chose four control cases and four cases in which we introduced a false finding, providing us with a total of eight cases. After making these alterations, we had 48 total cases, each of which had an associated CXR image and an impression section that was either unaltered (control) or altered according to the above process.

To generate the model outputs for evaluation, we first ran *ELIXR-B* phase 1 inference on the image to produce a report with ELIXR's findings. We then fed the image, the generated report, and a series of prompts that covered the main possible set of findings in chest x-rays to *ELIXR-B* phase 2.

For each prompt, the Q-Former-encoded image embeddings preceded the text hard prompt as input into PaLM 2-S, and the control or altered impressions section was inlined in place of the variable {altered_impression}. ELIXR provided an assessment as to whether each finding existed, and if so, assessed the laterality in order to support the laterality swap detection task. The prompts were as follows.

```
1. If there's an endotracheal tube (ET tube) in the chest x-ray, tell me
   whether it's mal-positioned or well-positioned. If there's no ET tube,
   respond 'no'
2. Is there any evidence of pneumothorax in the chest x-ray? If so, on which
   side(s)?
3. Are there any signs of pleural effusion present in the x-ray? If so, on
   which side(s)?
4. Are there any visible signs of pulmonary edema in the CXR? If so, on which
   side(s)?
5. Are there any signs of pneumonia or lung infection? If so, on which
   side(s)?
6. Are there any signs of consolidation or lung infection in this patient's
   chest x-ray? If so, on which side(s)?
7. Are there any signs of atelectasis in the lungs? If so, on which side(s)?
8. Are there any signs of fibrosis in the lungs? If so, describe it
9. Are there signs suggestive of a nodule or mass in this patient's chest
   x-ray? If so, on which side(s)?
10.Is the cardiac silhouette size normal or enlarged?
11.Is a hiatal hernia present? If so, on which side(s)?
12.Are there any signs of acute skeletal fracture? If so, where?
```

This initial part of the workflow was essentially comprehensive VQA. We concatenated these responses into a single piece of text to constitute ELIXR's comprehensive findings.

We then fed the concatenated questions and answers into a non-image-aligned LLM, Med-PaLM 2[29], to determine whether there were any missing findings, erroneously added findings, or laterality swaps. The two prompts to do this were as follows. Note that while MIMIC-CXR contains full reports for most cases, we altered and evaluated only the impression section of the reports.

```
You are an expert radiologist. These are your responses to a comprehensive
assessment of a patient's chest x-ray (CXR).
```

```
ASSESSMENT: {questions and ELIXR answers separated by new lines}.

A radiology resident has written the following radiology report for the same
CXR. RESIDENT'S REPORT: {altered_impression}.

Are there any findings that you mark positive or abnormal in your assessment but
that the resident either marks absent/negative or simply does not mention? If
so, what are the findings?
```

```
You are an expert radiologist. These are your responses to a comprehensive
assessment of a patient's chest x-ray (CXR).

ASSESSMENT: {questions and ELIXR answers separated by new lines}.

A radiology resident has written the following radiology report for the same
CXR. RESIDENT'S REPORT: {altered_impression}.

Are there any findings that you mark negative or normal in your assessment but
that the resident marks positive/abnormal in his report? If so, what are the
findings?
```

LLM responses were graded by a board-certified thoracic radiologist (CL) according to the rubric in Supplementary table 7. If the LLM correctly described the alteration (or identified an unaltered, control report as being correct and complete), the LLM output was scored as correct; if the LLM failed to identify the alteration, even if it gave an otherwise correct response, the output was scored as incorrect.

# Results

## ELIXR demonstrates state-of-the-art zero-shot classification performance comparable to fully supervised SupCon classifiers trained on as many as 224,000 examples

*ELIXR-B* and *ELIXR-C* demonstrated zero-shot classification performance on five findings ("atelectasis", "cardiomegaly", "consolidation", "pleural effusion", and "pulmonary edema") that was comparable to SupCon's classification performance when trained on the entirety of the CheXpert train set (~224,000 examples). Note that although ELIXR's vision encoder was initialized from a SupCon checkpoint, CheXpert was not used for pretraining SupCon at all, only for training small downstream classifiers. Thus, CheXpert is completely held out from *ELIXR-B* and *ELIXR-C* zero-shot.

Across the 13 findings (excluding "No Finding") from the CheXpert test set, *ELIXR-C* and *ELIXR-B* both surpassed the state-of-the-art zero-shot performance from CheXzero[23]. Figure 2 shows the details of the performance comparison for zero-shot classification. Positive and negative texts for prompt tuning in zero-shot classification are listed in Supplementary table 1.

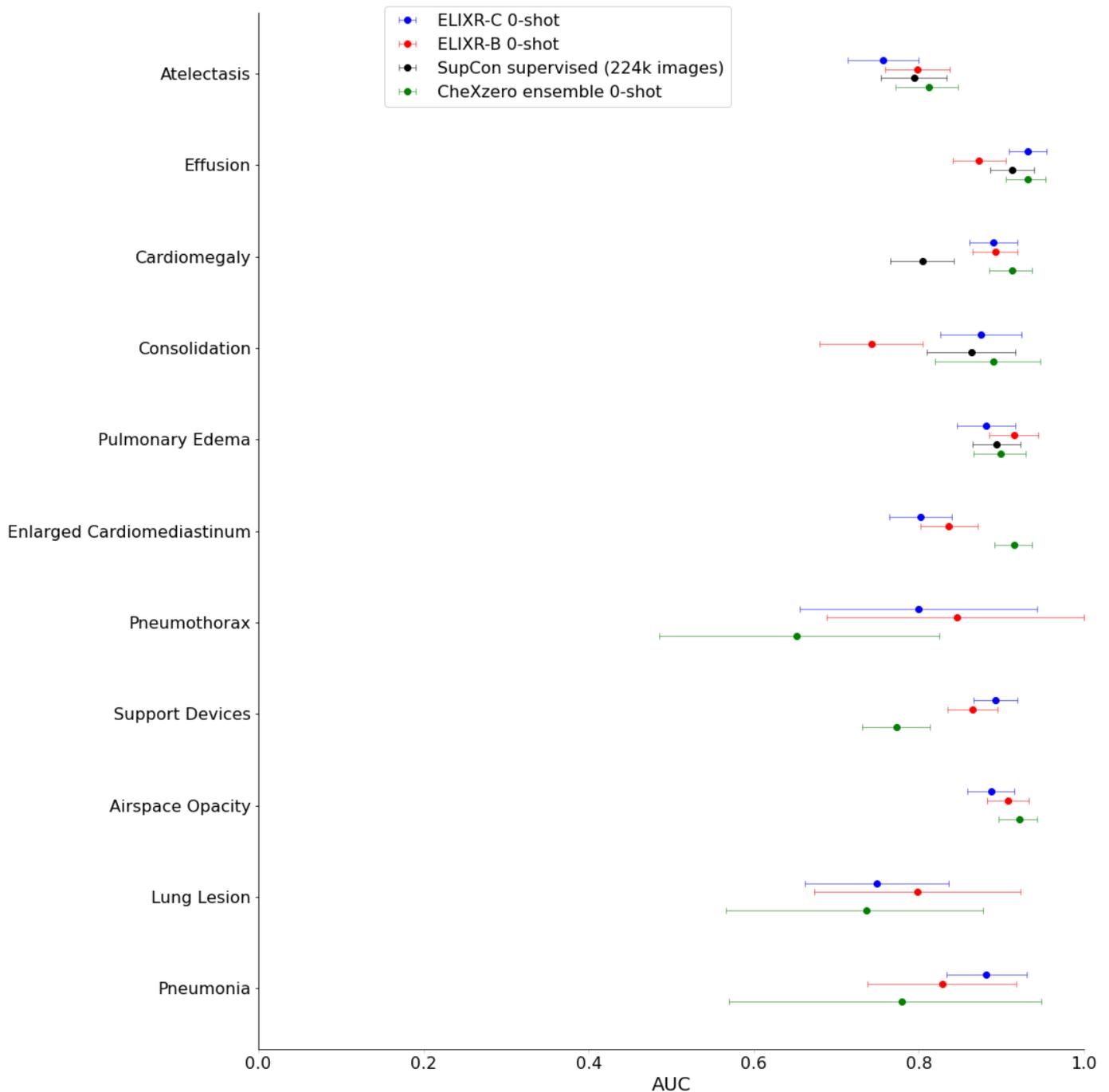

**Figure 2: ELIXR demonstrated state-of-the-art zero-shot classification performance comparable to label-efficient method supervised contrastive learning (SupCon).** AUCs and 95% confidence intervals for zero-shot classification for *ELIXR-B*, *ELIXR-C*, and CheXzero ensemble[23] across 11 findings (including only findings with >5 positives and excluding "No Finding" label) as well as SupCon fully-supervised (trained on the entire 224K examples of CheXpert) classification for five findings on the CheXpert test dataset. Full results with p-values are available in Supplementary table 8.

*ELIXR-B* and *ELIXR-C* both set a new state-of-the-art for data-efficient linear probe classification on CheXpert's five main findings ("atelectasis", "cardiomegaly", "consolidation", "pleural effusion", "pulmonary edema") using 1% and 10% of the train set, outperforming even the fully-finetuned ConVIRT[22]. *ELIXR-B* and *ELIXR-C* also both demonstrated data-efficient performance superior to SupCon (Figure 3) or, to put it another way, demonstrated data-efficient performance equivalent to SupCon using roughly two orders of magnitude less training data (e.g. *ELIXR-B* and *ELIXR-C* 64-shot performance was noninferior to SupCon 4096-shot performance; see Supplementary tables 9 and 10 for p-values). Table 3 shows a summary of comparisons between ELIXR and the SOTA for zero-shot and data-efficient.

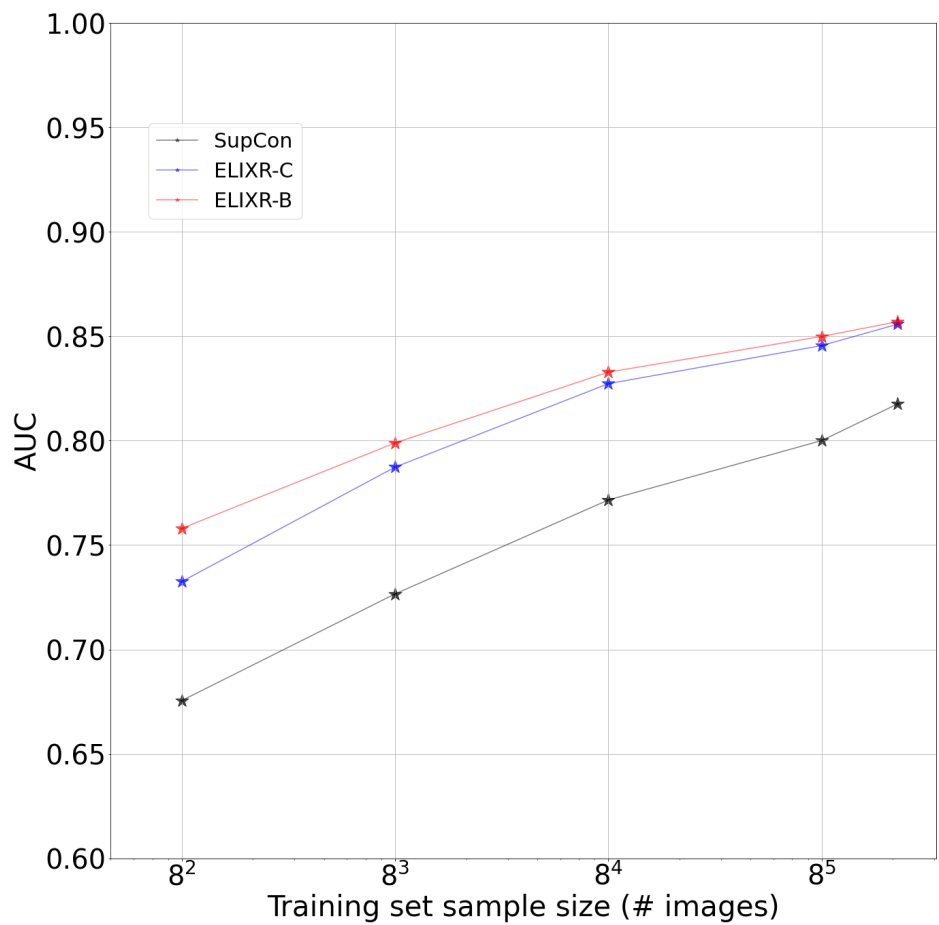

**Figure 3: Effect of using *ELIXR-C*, *ELIXR-B*, and supervised contrastive learning (SupCon) for data-efficient classification.** The reported performance is averaged across 2 datasets (CheXpert and Chest X-ray14) and seven findings: atelectasis, cardiomegaly, airspace opacity, fracture, pneumothorax, consolidation, pleural effusion, and pulmonary edema. Both *ELIXR-C* and *ELIXR-B* demonstrate superior performance compared to SupCon at matching dataset sizes, or, put another way, demonstrate performance on par with SupCon with two orders of magnitude less data (red and blue lines are translated two grid lines to the left from the black line). Detailed per-dataset and per-finding graphs are available in Supplementary figure 2. Delong's test results are available in Supplementary tables 9, 10.

|  | Mean AUC CheXpert test (5 main findings) | Mean AUC CheXpert test (13 findings) |
|---|---|---|
| **Zero shot** | | |
| CheXzero | **0.889** | 0.838 |
| *ELIXR-C* | 0.851 | **0.850** |
| *ELIXR-B* | 0.837 | 0.846 |
| **1% training data** | | |
| ConVIRT linear | 0.859 | -- |
| ConVIRT finetune | 0.870 | -- |
| *ELIXR-C* linear | 0.887 | -- |
| *ELIXR-B* linear | **0.893** | -- |
| **10% training data** | | |
| ConVIRT linear | 0.868 | -- |
| ConVIRT finetune | 0.881 | -- |
| *ELIXR-C* linear | 0.889 | -- |
| *ELIXR-B* linear | **0.898** | -- |

**Table 3: Comparison of *ELIXR* against state-of-the-art models, ConVIRT and CheXzero[22,23].** ELIXR sets a new state of the art, as measured by mean AUC, for zero-shot classification of 13 findings in CheXpert and data-efficient classification (1% and 10% training data) of 5 main findings in CheXpert.

## ELIXR enables state-of-the-art semantic search for findings using laterality-specific, severity-based, and nuanced terminology

*ELIXR-B* outperformed both *ELIXR-C* and the state-of-the-art MedCLIP[25] on the retrieval quality of top-5 retrieved images. For each query group, we computed the average metrics across the queries. NDCG@5 and Precision@5 of *ELIXR-B* were consistently better than *ELIXR-C* across all query groups. *ELIXR-B* scored higher than MedCLIP on NDCG@5 for all query groups and on Precision@5 (score = 2) for three out of four query groups (Table 4).

|  |  | Precision@5 (score=2) | Precision@5 (score=1) | NDCG@5 |
|---|---|---|---|---|
| **Findings** | MedCLIP | 0.29 | 0.29 | 0.25 |
|  | *ELIXR-C* | 0.63 | 0.66 | 0.66 |
|  | *ELIXR-B* | **0.74** | **0.74** | **0.74** |
| **Laterality** | MedCLIP | 0.77 | **1** | 0.76 |
|  | *ELIXR-C* | 0.73 | 0.8 | 0.83 |
|  | *ELIXR-B* | **0.93** | 0.93 | **0.94** |
| **Severity** | MedCLIP | **0.63** | **0.75** | 0.66 |
|  | *ELIXR-C* | 0.35 | 0.7 | 0.53 |
|  | *ELIXR-B* | 0.55 | 0.7 | **0.68** |
| **Nuanced** | MedCLIP | 0.54 | 0.68 | 0.54 |
|  | *ELIXR-C* | 0.6 | **0.84** | 0.73 |
|  | *ELIXR-B* | **0.64** | **0.84** | **0.74** |
| **Total** | MedCLIP | 0.50 [0.34-0.64] | 0.60 [0.40-0.77] | 0.49 [0.31-0.63] |
|  | *ELIXR-C* | 0.66 [0.52-0.78], 0.154 | 0.74 [0.60-0.86], p=0.296 | 0.68 [0.53-0.81], p=0.0912 |
|  | *ELIXR-B* | **0.75 [0.57-0.88], p=0.047** | 0.79 [0.63-0.91], p=0.143 | **0.76 [0.59-0.89], p=0.0234** |

**Table 4: Quantitative analysis of CXR semantic search using *ELIXR-C* and *ELIXR-B*.** *ELIXR-C* demonstrated the highest NDCG@5 scores across all four query groups and the highest Precision scores across three of four query groups as compared to *ELIXR-B* and the state-of-the-art MedCLIP. Confidence intervals were calculated from bootstrapping; p-values were calculated from permutation tests between *ELIXR-B* and MedCLIP or *ELIXR-C* and MedCLIP.

On twelve out of nineteen queries, *ELIXR-B* demonstrated perfect retrieval, including for laterality-specific queries like "right pleural effusion," severity-specific queries like "moderate cardiomegaly," and nuanced queries like "nasogastric tip reaches stomach." Notably, we found both *ELIXR-C* and *ELIXR-B* performed worse on fracture- and pneumothorax-related queries than queries related to other findings. This could be partially due to the low prevalence of these two pathologies in the MIMIC test set (4% for fracture and 3% for pneumothorax) compared to >10% prevalence for other pathologies. See Supplementary table 2 for a complete list of per-query scores.

### Finding (Both Correct)

Show me examples of central venous catheter

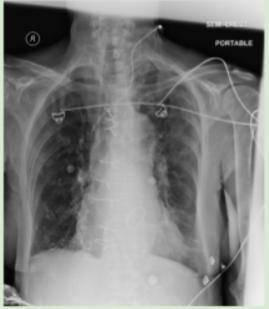

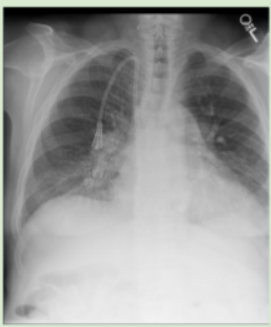

### Laterality (Both Correct)

Show me examples of left pneumonia

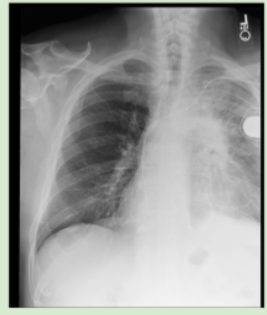

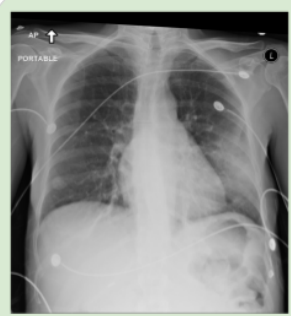

### Nuanced (Both Correct)

Show me examples of small right pleural effusion, no left pleural effusion

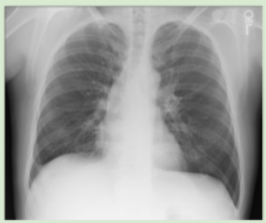

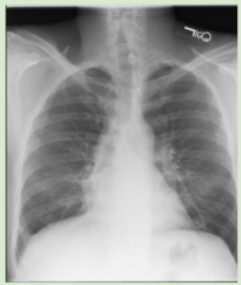

### Finding (Both Incorrect)

Show me examples of fracture

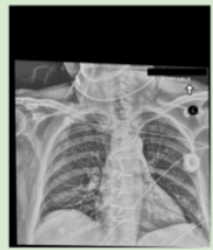

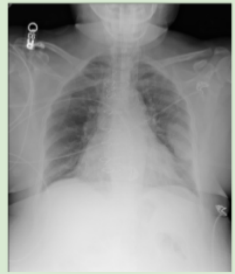
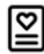

**Figure 4: Demonstration of semantic search using ELIXR.** Three of four examples here (top left, top right, bottom left) are correct for both images (scores of 2) while one example is incorrect for both images (scores of 0, bottom right).

In some cases, retrieval improved when using more specific queries, e.g. adding a laterality or severity modifier to a general finding. For example, *ELIXR-C* and *ELIXR-B* scored 0.723 and 0.83 for "left pneumothorax" as compared to 0.131 and 0.214 for "pneumothorax." These results point to the sensitivity of these models to prompting style.

## ELIXR supports visual question answering and quality assurance for radiology reports

On more challenging text-generation tasks, *ELIXR-B* demonstrated overall accuracies of 58.7% and 54.8% on two visual question answering (VQA) datasets, VQA-RAD (CXR-only questions) (Table 5) and MIMIC-CXR test (Table 6), as well as 62.5% on report quality assurance (QA) on MIMIC-CXR test (Tables 7, 8). Notably, *ELIXR-B* surpassed the accuracy of the SOTA model MedVInT which wasn't finetuned on VQA-RAD. A summary of VQA results appears in Tables 5 and 6. Quality assurance results appear stratified by alteration type in Table 7 and by primary finding in Table 8. Figure 5 shows a selection of example cases for both visual question answering and quality assurance.

| Answer type (A_TYPE) | Med-VInT w/ finetuning Accuracy** | Med-VInT w/o finetuning Accuracy | *ELIXR-B* Accuracy | *ELIXR-B* Sensitivity* | *ELIXR-B* Specificity* |
|---|---|---|---|---|---|
| both | **81.6% (451)** | 27.9% (574) | 58.7% (574) | N/A | N/A |
| closed | **86.8% (272)** | 28.2% (379) | 69.3% (379) | 42.6% (222) | 87.8% (195) |
| open | **73.7% (179)** | 27.1% (195) | 37.9% (195) | N/A | N/A |

**Table 5: VQA results of *ELIXR-B* on our VQA-RAD test set using semantic matching.** Number of total questions and answers in brackets. For comparison with the SOTA method, we provide results from MedVInT[30] both before and after finetuning on VQA-RAD to show the benefits it provides. *On the subset of expected that could be programmatically mapped to "yes" or "no". **Results from MedVInT for VQA-RAD finetuning are on all image modalities (not just chest X-ray) and from the official test split.

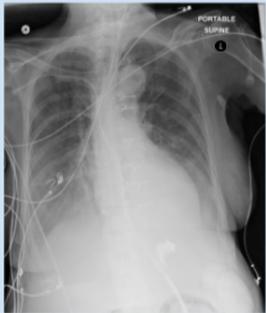
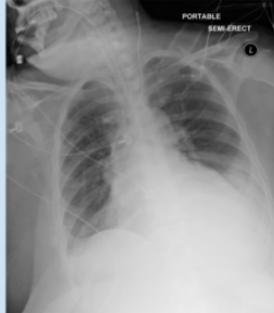
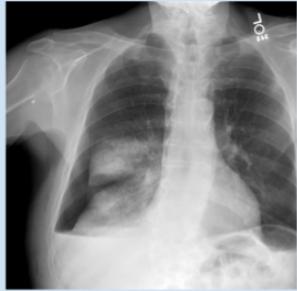
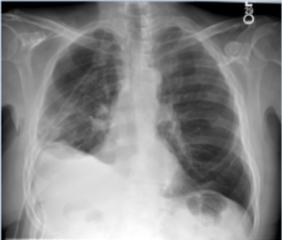
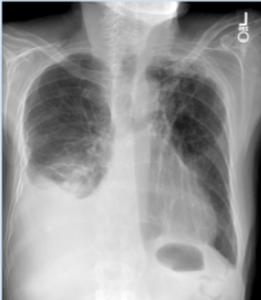
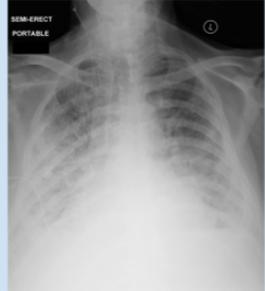

Figure 5: Qualitative results for visual question answering and quality assurance inference (from MIMIC test dataset).

| Question type | Accuracy |
|---|---|
| all | 54.8% (217) |
| presence | 64.5% (148) |
| location | 41.0% (39) |
| size, severity or type | 25.0% (30) |

**Table 6: Accuracy of *ELIXR-B*'s VQA answers on a subset of 48 MIMIC cases using expert-graded semantic matches.** Number of questions & answers noted in brackets. Nine questions were marked as non-gradable by the expert due to insufficient information or non-relevance (e.g. question about severity despite condition not being present).

| Alteration type | Number of cases | Overall model score (percent correct) |
|---|---|---|
| Control | 15 | 53.3% |
| Swap laterality | 8 | 87.5% |
| Add major finding | 15 | 60.0% |
| Remove major finding | 9 | 50.0% |
| **Total** | **48** | **60.4%** |

**Table 7: Summary statistics of report quality assurance results, stratified by alteration type.**

| Primary finding | Number of cases | Overall model score (percent correct) |
|---|---|---|
| No finding | 8 | 100% |
| Pneumothorax | 8 | 25% |
| Pleural Effusion | 8 | 62.5% |
| Edema | 8 | 62.5% |
| Consolidation or Pneumonia | 8 | 75% |
| Lung Lesion | 8 | 37.5% |
| **Total** | **48** | **60.4%** |

**Table 8: Summary statistics of quality assurance results, stratified by primary finding.**

It is important to note that the results we report on VQA-RAD are not directly comparable to those reported in the literature, for the following reasons: (1) we only used a CXR subset of VQA-RAD, since ELIXR currently is limited to this modality, (2) we opted for a more difficult split than the official development/test split in which the same image never appears across splits, (3) we refrained from training on the development set and only used it for checkpoint selection. The effect of (2) and (3) appears to be large: as Table 5 shows, MedVInT's performance increases from 27.1% to 73.7% on open-ended accuracy and from 28.2% to 86.8% on close-ended accuracy after finetuning on VQA-RAD. Moor and Huang et al[31] noted this data leakage in the official VQA-RAD splits, as well.

## Discussion

In this study, we developed and evaluated ELIXR, a multimodal model that grafts a language-aligned vision encoder onto a frozen LLM. The model was trained using CXR images paired with their free-text radiology reports, without the requirement for expensive expert data curation. The model achieved state-of-the-art performance for zero-shot classification, data-efficient classification, and semantic search tasks, while also demonstrating potential in visual question answering and radiology report quality assurance. The modular architecture has the advantage of being easily adaptable to other tasks, with the ability to swap in different vision encoders and base LLMs as required.

The ELIXR architecture is data and computationally efficient. Previously we demonstrated that small nonlinear classifiers trained on a frozen SupCon vision encoder can outperform fully supervised models in low-data settings[12]. With ELIXR, we have improved upon the data efficiency of supervised contrastive learning by two orders of magnitude. This offers the potential to train highly accurate models that are capable of addressing the long tail of diagnoses (including rare diseases), with only a fraction of the requirement for expert-curated training data. The process of prototyping a model for a new task also becomes simplified and more widely accessible, requiring only the design of positive and negative prompts using natural language without a requirement for machine learning expertise. We hope that these approaches will enable a wider range of researchers to engage in a broader array of diagnostic research questions and will allow medical professionals to develop models for underserved populations or understudied diseases.

By leveraging a pre-trained frozen vision encoder and a frozen LLM, we were able to train ELIXR in a highly compute-efficient manner. Backpropagation of gradients during the second stage of training is only performed for the Q-Former and MLP components, which are orders of magnitude smaller than the vision encoder and frozen LLM. In comparison to finetuning the vision encoder and/or LLM, this approach to training can be easily used by others who do not have access to substantial compute hardware. Furthermore, when adapting to a new task, or when newer generations of vision encoders and LLMs become available, it is straightforward and relatively inexpensive to train a new ELIXR model to take advantage of these advances.

LLMs offer a deep natural language understanding that enables a new range of possibilities for rich interaction between clinicians and AI. In this study, we demonstrated early promising capabilities in semantic search, vision question answering, and report quality assurance–all tasks that cannot be achieved easily using traditional vision-only models.

Semantic search unlocks the ability to search within an image for features of interest using free text prompts. ELIXR demonstrated high retrieval precision across a broad range of queries, including perfect retrieval of five out of seven findings-related queries (e.g. "central venous catheter"), two out of three laterality-specific queries (e.g. "right pleural effusion"), two out of four severity-related queries (e.g. "moderate cardiomegaly"), and three out of five nuanced queries (e.g. "nasogastric tube reaches stomach"). This capability could be used by researchers to identify images for study datasets, by clinicians to search for specific images of interest from a patient's historical record, by educators to find examples for teaching purposes, as well as in many other applications.

ELIXR also enables rich human-AI interaction through visual question answering. We benchmark our VQA performance on the chest X-ray subset of the VQA-RAD dataset, yielding accuracy of 69.1% across closed type questions, and 37.9% across open type questions. In contrast to others' work, we do not train on VQA-RAD, as the standard VQA-RAD data splits exhibit an overlap with images in both train/test that risk artificially inflating performance[31]. In addition, we report accuracy based on assessment by a board-certified thoracic radiologist rather than unigram matching (BLEU-1), since BLEU-1 does not comprehensively reflect the quality of VQA.

Finally, we demonstrate the ability of ELIXR to use its understanding of CXRs to check for errors in written radiology reports. We envision that such a capability could have utility in the hospital setting, potentially acting as an advanced multimodal "spell check" to alert radiologists to suspected inconsistencies in their reports, improving both quality and consistency of care. It could also act as an "AI Mentor", evaluating and mentoring more junior radiologists, including in settings where resource limitations result in less opportunity for oversight by senior radiologists. VQA and QA were heavily dependent on the specific prompts used, which were in turn affected by the LLM used by ELIXR. With further work into prompt engineering, it is expected that one could make the QA more specific, with fewer false positive reports. This high specificity is likely to be needed in a clinical environment, especially where the error rate is far lower than in this enriched simulated set.

There are a number of limitations to our work. Firstly, through employing a frozen LLM and vision encoder, we achieve our goals of training efficiency, but this might be at the expense of overall performance (although some works suggest otherwise[32]). Secondly, ELIXR inherits the current set of wider challenges of LLMs including fragility to changes in prompts, hallucinations, and unwarranted confidence in its answers, even when wrong[33]. We expect that advances in future generations of LLMs will directly lead to improved performance when incorporated into ELIXR. Thirdly, a lack of established, robust benchmarks makes it challenging to compare performance and establish state-of-the-art. Fourthly, we note that the MIMIC CXRs used were highly complex in an intensive care setting, containing multiple findings, which added more complexity than is typical compared to routine hospital X-rays. A non-intensive care CXR dataset that reflects routine wider hospital practice would be valuable to evaluate in future work. Finally, we selected different stages of ELIXR for different tasks based upon a performance versus efficiency trade-off. For example, on zero-shot learning tasks, the benefits of *ELIXR-B* over *ELIXR-C* were limited. In contrast, VQA and report QA tasks are more text dependent, and benefited from longer training using larger amounts of text data. We found that some categories of findings were challenging across all tasks–for example pneumothorax and fractures–where the difficulty of these cases is only partly explained by their low prevalence and noisy reference standards in the training datasets[5].

In future work, we hope to explore ELIXR's performance with different general purpose and medically specialized LLMs (e.g. Med-PaLM 2). We are also excited by the possibility of extending these methods to other imaging modalities such as musculoskeletal X-ray, mammography, computed tomography (CT), and also beyond radiology in an effort we are calling Medical Information Adapters. It is also likely that temporal information can be incorporated into this architecture, widening the range of potential real world clinical applications.

# Conclusion

In this study, we developed and evaluated an efficient vision-language multimodal model for medical imaging that is trained using only medical images paired with free-text radiology reports obtained from routine clinical practice. The method is highly compute and data efficient to train, and we demonstrated promising performance across a range of multimodal radiology tasks. This work is an initial step towards a general purpose X-ray artificial intelligence system.

# Acknowledgements

We acknowledge Yuan Liu and Karan Singhal from Google Research for their critical feedback on the manuscript. We thank the NIH Clinical Center for making the ChestX-ray14 dataset publicly available, the MIT Laboratory for Computational Physiology for making the MIMIC-CXR dataset publicly available, and the Stanford ML Group for making the CheXpert dataset publicly available. We also thank the Google Research team for software and hardware infrastructure support.

# Author contributions

S.X., L.Y., S.S., D.T., S.P., C.K., D.G., R.P., A.S. contributed to the conception of the study and study design; M.S., P.S., C.C., C.K., D.G., R.P., A.S. contributed to acquisition of the data; S.X., L.Y., M.S., T.K., M.M., W-H.W., A.K., S.K., Z.M., Y.L., C.L., S.P., C.K., D.G., R.P., A.S. contributed to analysis and interpretation of the data; P.S., Y.L., P.S., M.E., S.R.K., Y.M., K.C., G.C., S.S., D.T., K.E. provided strategic guidance; L.Y., W-H.W., P.S., Y.L., S.P., C.K., D.G., R.P., A.S. contributed to paper organisation and team logistics; S.X., L.Y., M.S., T.K., M.M., W-H.W., A.K., S.K., Z.M., Y.L., D.T., C.K., D.G., R.P., A.S. contributed to drafting and revising the manuscript.

# Supplementary materials

**Supplementary text**

- Additional ELIXR training details

**Supplementary tables**

- Supplementary table 1: Prompts used for zero-shot classification. We adopt the prompting strategy developed by CheXzero (Tiu et al. 2022).
- Supplementary table 2: Queries used for semantic search and the ELIXR performance using normalized discounted cumulative gain (NDCG).
- Supplementary table 3: Prompts used for MIMIC-CXR ground truthing.
- Supplementary table 4: Questions used for visual question answering (VQA) evaluation on MIMIC-CXR test cases.
- Supplementary table 5: Visual question answering evaluation scoring rubric.
- Supplementary table 6: Methodology for altering impressions for report quality assurance evaluation.
- Supplementary table 7: Scoring rubric for quality assurance large language model (LLM) output.
- Supplementary table 8: Zero-shot classification using *ELIXR-C* and *ELIXR-B* across 13 findings in the CheXpert test set.
- Supplementary table 9: Data-efficient classification performance comparison between *ELIXR-B* and *ELIXR-C*.
- Supplementary table 10: Data-efficient classification performance comparison between *ELIXR-B* and supervised contrastive learning (SupCon) by sample size.

**Supplementary figures**

- Supplementary figure 1: Suspect positive or negative ground truth MIMIC labels as identified by the query to Med-PaLM 2 truthed by a board certified radiologist.
- Supplementary figure 2: Data-efficient classification performance between models per-dataset and per-finding using different training set sample sizes.

# Supplementary Text

## Additional ELIXR training details

### ELIXR-C

For the first stage of training using the CLIP loss, EfficientNet-L2[34] was used as the vision encoder, initialized from the SupCon checkpoint used in Sellergren et al[12]. The small variant of T5 was used for the text encoder, initialized from scratch[13]. The T5 text encoder used a SentencePiece model[35] with a vocabulary size of 32,000 pretrained on PubMed abstracts. For data augmentation, horizontal flipping and random rotation up to 15 degrees were applied. Images were resized to 1280x1280 pixels. A dimension of 128 for the projection head and a fixed temperature of 0.07 were used. Preprocessing on the radiology reports was done to select, in order of preference, the impression section, followed by the findings section. The input text sequence was truncated at 128 tokens. A batch size of 64 was split across 128 TPUv3 cores using spatial partitioning. Stochastic gradient descent (SGD) constant learning rate was set to 0.0001 with a momentum of 0.98. The model was trained for roughly 80,000 steps. Training was done in TensorFlow.

### ELIXR-B

In the first phase of *ELIXR-B* training, the embeddings from *ELIXR-C* were precomputed for all datasets and spatially pooled from 40x40x1376 to 8x8x1376. We used 32 BLIP-2 query tokens and a max text sequence length of 128. Q-Former weights were initialized from BERT-base. A batch size of 128 was split across 8 TPUv3 cores. Constant learning rate of 1e-5 was used with the Adam optimizer (beta1 of 0.98, beta2 of 0.999, epsilon of 1e-8). Training was done in Jax. Checkpoints were selected using zero-shot performance on CheXpert validation as well as the validation losses. More specifically, the checkpoint that performs the best on zero-shot AUC, image-text matching loss, and contrastive loss was used for zero/data-efficient and semantic search, while the checkpoint that performs the best on image-grounded text generation was used for initial checkpoint for phase 2 training.

In the second phase of *ELIXR-B* training, 32 replicas of PaLM2 S were launched as inference servers, each on 8 TPUv3 cores. When called for given inputs (the output query tokens from the Q-Former plus any additional text tokens), the server provides back the gradients (based on the language modeling loss) which can be backpropagated through the Q-Former. For this phase, a batch size of 32 was used, 1 per inference server replica. Constant learning rate of 5e-4 was used with the Adam optimizer (beta1 of 0.98, beta2 of 0.999, epsilon of 1e-8).

# Supplementary tables

**Supplementary table 1: Prompts used for zero-shot classification. We adopt the prompting strategy developed by CheXzero[23].**

| Condition | Positive prompts | Negative prompts |
|---|---|---|
| Enlarged cardiomediastinum | widened cardiomediastinum | no acute cardiopulmonary process<br>cardiomediastinal silhouette is normal |
| Cardiomegaly | mild cardiomegaly<br>moderate cardiomegaly<br>severe cardiomegaly | heart size is normal<br>no acute cardiopulmonary process<br>normal study |
| Lung lesion | lytic lesion<br>cavitary lesion<br>parenchymal lesion | no acute cardiopulmonary process |
| Airspace opacity | bilateral opacities<br>basal opacity | no focal opacity<br>lung volumes are normal |
| Edema | mild pulmonary edema<br>moderate pulmonary edema<br>severe pulmonary edema | no pulmonary edema<br>no acute cardiopulmonary process<br>normal study |
| Consolidation | suggestive of consolidation | normal study |
| Pneumonia | suggestive of pneumonia | lungs are clear<br>no acute cardiopulmonary process |
| Atelectasis | plate atelectasis<br>subsegmental atelectasis | lungs are clear<br>no acute cardiopulmonary process<br>normal study |
| Pneumothorax | apical pneumothorax | no pneumothorax |
| Pleural effusion | left pleural effusion<br>right pleural effusion<br>bilateral pleural effusions | no acute cardiopulmonary process<br>normal study |
| Pleural other | blunting of costophrenic angle<br>pleural thickening | no acute cardiopulmonary process |
| Fracture | rib fractures | no acute cardiopulmonary process |
| Support Devices | monitoring and support devices<br>NG tube<br>ET tube<br>catheter<br>PIC line | no acute cardiopulmonary process |

**Supplementary table 2: Queries used for semantic search and the ELIXR performance using normalized discounted cumulative gain at five (NDCG@5).**

| Topic | Queries | NDCG@5 ELIXR-C | NDCG@5 ELIXR-B | NDCG@5 MedCLIP |
|---|---|---|---|---|
| Single finding | Pneumothorax | 0.131 | 0.214 | 0.0 |
| | Pneumonia | 0.927 | 1.0 | 0.0 |
| | Central venous catheter | 1.0 | 1.0 | 0.447 |
| | Nasogastric tube | 0.869 | 1.0 | 0.699 |
| | Endotracheal tube | 1.0 | 1.0 | 0.146 |
| | Fracture | 0 | 0 | 0.301 |
| | Pacemaker | 0.723 | 1.0 | 0.146 |
| Laterality | Left Pneumothorax | 0.723 | 0.83 | 0.5 |
| | Right pleural effusion | 1.0 | 1.0 | 1.0 |
| | Left Pneumonia | 0.769 | 1.0 | 0.765 |
| Severity | Small pleural effusion | 0.616 | 0.446 | 0.777 |
| | Large Pneumothorax | 0.066 | 0.254 | 0.0 |
| | Moderate Pulmonary edema | 0.5 | 1.0 | 0.927 |
| | Moderate Cardiomegaly | 0.927 | 1.0 | 0.934 |
| Nuanced features | Small right pleural effusion, no left pleural effusion | 0.68 | 1.0 | 0 |
| | Mild right Pneumonia | 0.786 | 0.449 | 0.764 |
| | Loculated pleural effusion | 0.478 | 0.246 | 0.488 |
| | Multiple focal lung opacities | 1.0 | 1.0 | 0.746 |
| | Nasogastric tube reaches stomach | 0.701 | 1.0 | 0.699 |

**Supplementary table 3: Prompts used for MIMIC-CXR ground truthing.** Keywords for selecting candidates for MIMIC-CXR reports followed by prompts used to determine if the given condition exists. These were used to select candidates for confirmation by a board certified thoracic radiologist (CL).

| Condition | Key Words |
|---|---|
| atelectasis | atelectasis |
| cardiomegaly | cardiomegaly<br>cardiac silhouette |
| catheter | central venous catheter<br>central line<br>picc |
| fracture | fracture<br>acute fracture<br>fx |
| hilar enlargement | hilar enlargement |
| lung opacity | lung opacity<br>lung opacities<br>infiltrate<br>pneumonia<br>atelectasis<br>consolidation<br>airspace opacity<br>airspace opacities |
| mediastinal widening | abnormal mediastinal widening<br>widened mediastinum<br>widened |
| nodule | lung nodule<br>nodule<br>nodular opacity |
| pleural effusion | pleural effusion |
| pneumonia | pneumonia |
| pneumothorax | pneumothorax |
| pulmonary edema | pulmonary edema<br>pulmonary vascular congestion |
| tube | endotracheal tube<br>enteric tube<br>ng |

|  | og<br>feeding tube<br>et tube |
| --- | --- |
| **Prompt used with above text on Med-PaLM 2** ||
| All key words except cardiac silhouette ||
| You are a helpful medical knowledge assistant. Provide useful, complete, concise, and scientifically-grounded queries to radiology reports.<br>Does this report mention that the patient has a {keyword}? Report:{description} ||
| cardiac silhouette ||
| You are a helpful medical knowledge assistant. Provide useful, complete, concise, and scientifically-grounded queries to radiology reports.<br>Does this report mention that the patient has a cardiac silhouette that is enlarged? Report:{description} ||

**Supplementary table 4: Questions used for visual question answering evaluation on MIMIC-CXR test cases.** For each case, the model was queried with a findings-specific set of questions, depending on the corresponding label in the MIMIC CXR JPG dataset being set to 1.0, with questions covering the condition's presence, location, and severity, size or type. An additional two findings-independent questions were asked per case, covering the presence of other conditions. For cases with none of the listed five MIMIC labels being set to 1.0, three label-independent questions were asked.

| Finding | Questions the model was queried with | Question type |
|---|---|---|
| Pneumothorax | Is a pneumothorax **present** ?<br>Is a pneumothorax **present** in this image ? | Presence |
| | **Where** is pneumothorax present ? | Location |
| | What is the **size** of pneumothorax, if present ? | Size/Severity/Type |
| Pleural Effusion | Does the patient **have** a pleural effusion ?<br>Is a pleural effusion **present** in this image ? | Presence |
| | What is the **location** of the pleural effusion, if present ? | Location |
| | What is the **size** of pleural effusion, if present ? | Size/Severity/Type |
| Edema | Is a pulmonary edema **present** in this image ?<br>Is there **evidence** of a pulmonary edema ? | Presence |
| | **Where** is pulmonary edema present ? | Location |
| | What is the **severity** of the pulmonary edema, if present ?<br>What is the **type** of the pulmonary edema, if present ? | Size/Severity/Type |
| Consolidation OR Pneumonia | Is a lung consolidation or pneumonia **present** in this image ? | Presence |
| | What is the **location** of the lung consolidation or pneumonia, if present ? | Location |
| Lung Lesion | Are lung nodules or a mass **present**?<br>Does the patient **have** lung nodules or a mass ? | Presence |
| | **Where** are lung nodules or a mass located ? | Location |
| Finding-independent | What abnormalities are seen within the lungs ?<br>Does the patient have cardiomegaly ?<br>What is the pathology ?<br>Is atelectasis present in this image ?<br>Is a pneumothorax **present** ?<br>Is a pneumothorax **present** in this image ?<br>Does the patient **have** a pleural effusion ?<br>Is a pleural effusion **present** in this image ?<br>Is a pulmonary edema **present** in this image ?<br>Is there **evidence** of a pulmonary edema ?<br>Is a lung consolidation or pneumonia **present** in this image ?<br>Is pneumonia present ?<br>Are lung nodules or a mass **present**?<br>Does the patient **have** lung nodules or a mass ? | Presence |

|  | Does the patient have lung opacity ?<br>Is a radiopaque foreign body or pacemaker present ?<br>Are there any findings ?<br>Is a hiatal hernia present ?<br>Are there any signs of interstitial fibrosis in the lungs ?<br>Are there any notable abnormalities in the imaged upper abdomen ?<br>Is there evidence of an endotracheal tube ?<br>Is a central venous catheter present ?<br>Is there evidence of a nasogastric or orogastric tube ?<br>Is a hilar enlargement present in this image ?<br>Is a mediastinal widening present in this image ?<br>Does the patient have a skeletal fracture ? |  |
| --- | --- | --- |

**Supplementary table 5: Visual question answering evaluation scoring rubric.** Definition of correctness scores used for assessing model answers against expected answers (for VQA-RAD) or the CXR image and/or report (for MIMIC test set) in visual question answering by radiologist.

| Summary | Score Details | Score |
|---|---|---|
| Correct | LLM-provided answer either:<br>● is an exact match to the ground truth answer.<br>● contains additional and still accurate information above what the ground truth specifies<br>● is another diagnosis consistent with the image that also answers the question. e.g. the associated image shows two potential diagnoses and the predicted answer provides the diagnosis not present in the ground truth, and the question did not specify which diagnosis it was looking for.<br>● is a rephrasing of the ground truth answer such that a patient's diagnosis would not differ in treatment from the ground truth. e.g. a predicted answer of "lobe collapse" with ground truth of "pneumothorax" would be counted as a correct prediction. | 1.0 |
| Partially Correct | LLM-provided answer has correct information but is in flawed in one of the following ways:<br>● the answer omits some key details (e.g. laterality, location) asked by the question and present in the ground truth.<br>● the answer provides incorrect key details, but the underlying condition is still accurate. | 0.5 |
| Incorrect | LLM-provided answer is either:<br>● entirely inaccurate.<br>● does not respond to the question or appears to answer a different one.<br>● is missing an answer.<br>● is internally inconsistent. | 0.0 |
| Ambiguous | LLM-provided answer cannot be compared to the ground truth. e.g. a ground-truth answer of "unsure" or "ambiguous" does not provide enough information to compare against. | N/A |

**Supplementary table 6: Methodology for altering impressions for report quality assurance evaluation.** Here we describe the specific alterations that were made to cases associated with each finding for the quality assurance task.

| Alteration | Primary Finding | Alteration rule |
|---|---|---|
| No change (control) | All | No change |
| Change laterality | All except No Finding and Edema | If the laterality of the primary finding is left-sided, replace to indicate right-sided or vice versa. Do not change laterality of any other finding. Cases with bilateral primary findings are excluded from this alteration category. |
| Remove major finding | All except No Finding | Remove all references to the primary finding and indicate either that there is "no evidence" of the finding or that the finding "has resolved," whichever is more appropriate in the context of the remaining impression. Retain content that relates to other findings. If no findings remain, change impression to read, "No acute cardiopulmonary process." |
| Add major finding | No Finding | Add "Medium right pleural effusion." Remove any sentences indicating the report is normal/clear. |
| Add major finding | Pneumothorax | Add: "Malpositioned endotracheal tube." |
| Add major finding | Pleural Effusion | Add: "Approximately 1 cm nodule in mid right lung." |
| Add major finding | Edema | Add: "Several acute displaced rib fractures." |
| Add major finding | Consolidation or Pneumonia | Add: "Large left pleural effusion." |
| Add major finding | Lung Lesion | Add: "Moderate right pneumothorax." |

**Supplementary table 7: Scoring rubric for quality assurance large language model (LLM) output.** Here we show the detailed scoring rubric that was used by the board-certified thoracic radiologist (CL) to assess the LLM output.

| Impression type being analyzed | LLM assessment by board-certified thoracic radiologist (CL) | Score |
|---|---|---|
| Control (no change) | LLM suggests that impression is correct | 1 |
| Control (no change) | LLM suggests a problem with the impression where the suggestion is correct or possibly correct but not important for case management (e.g., "report failed to characterize heart size") | 1 |
| Control (no change) | LLM suggests a problem with the impression where the suggestion is correct and possibly important for case management | Exclude as this was not a valid control |
| Control (no change) | LLM suggests a problem with the impression where the suggestion is incorrect | 0 |
| Remove primary finding | LLM suggests that impression is correct | 0 |
| Remove primary finding | LLM does not identify the primary finding that was removed (even if it makes any other suggestion that is correct or possibly correct) | 0 |
| Remove primary finding | LLM identifies the primary finding that was removed | 1 |
| Adding incorrect finding | LLM suggests that impression is correct | 0 |
| Adding incorrect finding | LLM does not correctly identify the added finding (even if it makes any other suggestion that is correct or possibly correct) | 0 |
| Adding incorrect finding | LLM correctly identifies the added finding | 1 |
| Swap primary finding laterality | LLM suggests that impression is correct | 0 |
| Swap primary finding laterality | LLM response does not identify the problem with the laterality of the primary finding (even if it makes any other suggestion that is correct or possibly correct) | 0 |
| Swap primary finding laterality | LLM identifies a problem with the laterality of the primary finding | 1 |

**Supplementary table 8: Zero-shot classification using *ELIXR-C* and *ELIXR-B* across 13 findings in the CheXpert test set.** Area under curves (AUCs) with 95% confidence intervals are reported. the "No Finding" label is excluded.

|  | Zero-shot | | | |
|---|---|---|---|---|
|  | *ELIXR-C* AUC [95% CI] | *ELIXR-B* AUC [95% CI] | Difference *ELIXR-B* vs. *ELIXR-C* [95% CI], p-value | Largest SupCon data-efficient sample size for which *ELIXR-B* is noninferior [95% CI], p-value |
| **CheXpert** | | | | |
| Atelectasis | 0.754 [0.714-0.795] | 0.798 [0.759-0.838] | 0.042 [0.00, 0.084] p=0.05272 | 224316, -0.004 [-0.045, 0.036] p=0.84095 |
| Effusion | **0.930** [0.908-0.951] | 0.873 [0.841-0.903] | -0.059 [-0.086, -0.032] p=0.00001 | 4096, 0.020 [-0.006, 0.047] p=0.13340 |
| Cardiomegaly | 0.891 [0.862-0.919] | 0.892 [0.861-0.921] | 0.002 [-0.023, 0.026] p=0.88496 | 224316, -0.088 [-0.120, -0.056] p<1e-5 |
| Consolidation | **0.875** [0.819-0.922] | 0.742 [0.680-0.804] | -0.133 [-0.199, -0.067] p=0.00008 | 512, 0.015 [-0.067, 0.096] p=0.72400 |
| Pulmonary Edema | 0.880 [0.843-0.913] | **0.915** [0.881-0.942] | 0.033 [0.015, 0.052] p=0.00033 | 224316, -0.021 [-0.046, 0.005] p=0.10897 |
| Enlarged Cardiomediastinum | 0.800 [0.763-0.837] | **0.837** [0.803-0.869] | 0.035 [0.013, 0.056] p=0.00195 | N/A |
| Pleural Other | 0.729 [0.467-1.000] | 0.490 [0.151-0.780] | -0.239 [-0.622, 0.143] p=0.22031 | N/A |
| Pneumothorax | 0.800 [0.630-0.932] | 0.846 [0.656-0.980] | 0.047 [-0.067, 0.160] p=0.42131 | N/A |
| Support Devices | **0.894** [0.865-0.919] | 0.865 [0.835-0.892] | -0.027 [-0.047, -0.007] p=0.00712 | N/A |
| Airspace Opacity | 0.888 [0.857-0.915] | **0.908** [0.882-0.931] | 0.021 [0.000, 0.041] p=0.04913 | N/A |
| Lung Lesion | 0.747 [0.659-0.838] | 0.798 [0.671-0.914] | 0.049 [-0.064, 0.163] p=0.39248 | N/A |
| Pneumonia | 0.881 [0.833-0.930] | 0.828 [0.734-0.914] | -0.054 [-0.140, 0.033] p=0.22251 | N/A |
| Fracture | 0.637 [0.444-0.876] | 0.395 [0.173-0.679] | -0.242 [-0.637, 0.153] p=0.22950 | N/A |

**Supplementary table 9: Data-efficient classification performance comparison between *ELIXR-B* and *ELIXR-C*.** (a) CheXpert, (b) CXR-14. Delong's test results comparing data-efficient performance of *ELIXR-B* vs. *ELIXR-C* at matching dataset sizes for CheXpert and Chest X-ray14. Bold for *ELIXR-B*'s outperformance being statistically significant. Red if *ELIXR-C* outperformed *ELIXR-B*.

(a)

| CheXpert | Difference between *ELIXR-B* and *ELIXR-C* | | | | |
|---|---|---|---|---|---|
| Sample size | 64 | 512 | 4096 | 32768 | 224316 |
| Atelectasis | 0.057 [-0.006, 0.119] p=0.07509 | **0.045 [0.005, 0.085] p=0.02870** | 0.021 [-0.015, 0.057] p=0.24958 | 0.020 [-0.013, 0.053] p=0.23971 | -0.001 [-0.027, 0.026] p=0.95679 |
| Cardiomegaly | **0.098 [0.052, 0.144] p=0.00003** | **0.040 [0.006, 0.074] p=0.02128** | 0.032 [-0.001, 0.066] p=0.05489 | **0.019 [0.000, 0.038] p=0.04469** | 0.012 [-0.012, 0.035] p=0.32533 |
| Consolidation | 0.042 [-0.075, 0.159] p=0.48560 | 0.025 [-0.066, 0.117] p=0.59087 | -0.002 [-0.045, 0.041] p=0.91520 | 0.005 [-0.036, 0.047] p=0.79792 | 0.029 [-0.009, 0.068] p=0.13381 |
| Effusion | 0.001 [-0.023, 0.025] p=0.95793 | -0.001 [-0.020, 0.018] p=0.89187 | 0.002 [-0.013, 0.017] p=0.79330 | 0.007 [-0.009, 0.022] p=0.41349 | -0.004 [-0.017, 0.009] p=0.55608 |
| Pulmonary edema | -0.006 [-0.053, 0.041] p=0.81119 | 0.000 [-0.029, 0.029] p=0.99372 | 0.007 [-0.017, 0.031] p=0.57298 | 0.005 [-0.013, 0.022] p=0.61078 | 0.006 [-0.017, 0.029] p=0.59233 |

(b)

| CXR-14 | Difference between *ELIXR-B* and *ELIXR-C* | | | | | |
|---|---|---|---|---|---|---|
| Sample size | 64 | 512 | 4096 | 32768 | 68801 | 674533 |
| Airspace opacity | **0.035 [0.023, 0.047] p<1e-5** | **0.019 [0.010, 0.029] p=0.00007** | **0.016 [0.009, 0.022] p<1e-5** | **0.010 [0.004, 0.016] p=0.00086** | | -0.001 [-0.007, 0.005] p=0.77917 |
| Consolidation | **0.015 [0.001, 0.030] p=0.03916** | 0.010 [-0.001, 0.021] p=0.06217 | 0.005 [-0.001, 0.012] p=0.09394 | **0.006 [0.000, 0.013] p=0.04882** | 0.005 [-0.002, 0.013] p=0.12789 | |
| Effusion | <span style="color:red">-0.009 [-0.014, -0.003] p=0.00178</span> | -0.000 [-0.005, 0.005] p=0.99721 | -0.003 [-0.006, 0.001] p=0.15106 | -0.002 [-0.005, 0.000] p=0.10230 | <span style="color:red">-0.004 [-0.006, -0.001] p=0.01593</span> | |
| Fracture | 0.004 [-0.081, 0.090] p=0.92169 | -0.004 [-0.076, 0.068] p=0.91167 | 0.012 [-0.067, 0.091] p=0.76612 | -0.016 [-0.084, 0.052] p=0.64405 | | -0.035 [-0.099, 0.028] p=0.27689 |
| Pneumothorax | **0.067 [0.032, 0.102]** **p=0.00021** | -0.030 [-0.059, -0.001] p=0.04248 | -0.012 [-0.036, 0.012] p=0.31538 | 0.007 [-0.005, 0.020] p=0.24105 | | 0.007 [-0.006, 0.020] p=0.30800 |
| Pulmonary edema | -0.005 [-0.028, 0.017] p=0.63723 | **0.019 [0.006, 0.033]** **p=0.00459** | -0.007 [-0.016, 0.002] p=0.13283 | -0.001 [-0.007, 0.004] p=0.68002 | -0.000 [-0.006, 0.006] p=0.99304 | |

**Supplementary table 10: Data-efficient classification performance comparison between *ELIXR-B* and supervised contrastive learning (SupCon) by sample size.** (a) CheXpert, (b) CXR-14. AUC differences between SupCon and *ELIXR-B* pretraining by sample sizes. Values in each cell represent the difference in AUC, with 95% CIs in square braces, and p-values. Bold indicates where the *ELIXR-B* approach performs noninferior (at a margin of 0.05) to SupCon at a smaller sample size.

(a)

| CheXpert | Sample Size ELIXR-B | 64 | 512 | 4096 | 32768 | 224316 |
|---|---|---|---|---|---|---|
| | Sample size SupCon | | | | | |
| Atelectasis | 224316 | -0.048 [-0.094, -0.002] p=0.04094 | -0.051 [-0.088, -0.015] p=0.00613 | **-0.013 [-0.048, 0.023] p=0.47990** | **0.007 [-0.009, 0.022] p=0.39980** | |
| | 32768 | **-0.025 [-0.076, 0.026] p=0.33903** | **-0.028 [-0.073, 0.017] p=0.21884** | **0.010 [-0.035, 0.056] p=0.65029** | | 0.034 [0.002, 0.066] p=0.04019 |
| | 4096 | **-0.011 [-0.058, 0.037] p=0.66183** | **-0.014 [-0.054, 0.026] p=0.49807** | | 0.044 [0.012, 0.076] p=0.00665 | 0.048 [0.019, 0.077] p=0.00114 |
| | 512 | **0.039 [-0.003, 0.081] p=0.07230** | | 0.074 [0.034, 0.114] p=0.00032 | 0.093 [0.053, 0.134] p<1e-5 | 0.098 [0.057, 0.138] p<1e-5 |
| | 64 | | 0.121 [0.066, 0.177] p=0.00002 | 0.160 [0.108, 0.212] p<1e-5 | 0.179 [0.126, 0.232] p<1e-5 | 0.183 [0.132, 0.235] p<1e-5 |
| Cardiomegaly | 224316 | -0.057 [-0.111, -0.003] p=0.03936 | **-0.018 [-0.062, 0.027] p=0.43275** | **-0.011 [-0.047, 0.025] p=0.55874** | **-0.004 [-0.027, 0.020] p=0.74519** | |
| | 32768 | **-0.038 [-0.094, 0.017] p=0.17618** | **0.001 [-0.043, 0.045] p=0.97401** | **0.008 [-0.026, 0.042] p=0.65056** | | 0.029 [0.008, 0.051] p=0.00653 |
| | 4096 | **-0.018 [-0.066, 0.029] p=0.45020** | **0.021 [-0.013, 0.054] p=0.22596** | | 0.035 [-0.002, 0.071] p=0.06508 | 0.049 [0.016, 0.083] p=0.00371 |
| | 512 | **0.013 [-0.034, 0.061] p=0.58649** | | 0.059 [0.020, 0.099] p=0.00338 | 0.066 [0.022, 0.110] p=0.00317 | 0.081 [0.038, 0.124] p=0.00023 |
| | 64 | | 0.123 [0.068, 0.178] p=0.00001 | 0.130 [0.083, 0.177] p<1e-5 | 0.137 [0.082, 0.192] p<1e-5 | 0.152 [0.099, 0.205] p<1e-5 |
| Consolidation | 224316 | -0.095 [-0.179, -0.011] p=0.02671 | **-0.039 [-0.105, 0.027] p=0.25086** | **0.001 [-0.056, 0.058] p=0.97076** | **0.012 [-0.021, 0.044] p=0.48487** | |
| | 32768 | **-0.075 [-0.155, 0.006]** | **-0.018 [-0.084, 0.048]** | **0.021 [-0.034, 0.077]** | | 0.033 [-0.009, 0.074] p=0.12735 |

|  | Sample size | | | | | |
|---|---|---|---|---|---|---|
| Pleural effusion (cont.) | | p=0.06871 | p=0.58336 | p=0.44896 | | |
| | 4096 | **-0.058 [-0.146, 0.030] p=0.19550** | **-0.002 [-0.060, 0.056] p=0.94469** | | 0.048 [-0.007, 0.104] p=0.08743 | 0.049 [-0.000, 0.098] p=0.05150 |
| | 512 | **0.011 [-0.096, 0.117] p=0.84267** | | 0.107 [0.033, 0.180] p=0.00450 | 0.117 [0.036, 0.199] p=0.00494 | 0.118 [0.036, 0.200] p=0.00465 |
| | 64 | | 0.175 [0.081, 0.268] p=0.00026 | 0.214 [0.134, 0.295] p<1e-5 | 0.225 [0.139, 0.311] p<1e-5 | 0.226 [0.142, 0.309] p<1e-5 |
| Pleural effusion | 224316 | -0.078 [-0.115, -0.042] p=0.00002 | -0.044 [-0.070, -0.018] p=0.00095 | **0.002 [-0.019, 0.023] p=0.83881** | **0.001 [-0.014, 0.017] p=0.89296** | |
| | 32768 | -0.071 [-0.107, -0.035] p=0.00011 | -0.037 [-0.063, -0.011] p=0.00583 | **0.009 [-0.011, 0.030] p=0.36040** | | 0.012 [-0.001, 0.026] p=0.07923 |
| | 4096 | -0.061 [-0.093, -0.029] p=0.00021 | -0.026 [-0.048, -0.005] p=0.01582 | | 0.019 [0.004, 0.033] p=0.01237 | 0.022 [0.003, 0.042] p=0.02590 |
| | 512 | **-0.025 [-0.059, 0.010] p=0.16249** | | 0.056 [0.027, 0.084] p=0.00012 | 0.055 [0.027, 0.082] p=0.00009 | 0.059 [0.027, 0.091] p=0.00032 |
| | 64 | | 0.077 [0.038, 0.115] p=0.00009 | 0.123 [0.085, 0.161] p<1e-5 | 0.122 [0.082, 0.161] p<1e-5 | 0.126 [0.084, 0.167] p<1e-5 |
| Pulmonary edema | 224316 | -0.066 [-0.107, -0.026] p=0.00130 | -0.036 [-0.069, -0.003] p=0.03056 | **-0.011 [-0.031, 0.009] p=0.28688** | **-0.003 [-0.019, 0.014] p=0.74628** | |
| | 32768 | -0.048 [-0.087, -0.010] p=0.01442 | **-0.018 [-0.054, 0.018] p=0.33455** | **0.007 [-0.018, 0.032] p=0.57217** | | 0.029 [0.006, 0.051] p=0.01272 |
| | 4096 | **-0.032 [-0.068, 0.005] p=0.09393** | **-0.001 [-0.038, 0.036] p=0.94866** | | 0.032 [0.009, 0.055] p=0.00608 | 0.045 [0.018, 0.073] p=0.00114 |
| | 512 | **-0.006 [-0.040, 0.029] p=0.75042** | | 0.050 [0.018, 0.082] p=0.00200 | 0.058 [0.025, 0.091] p=0.00049 | 0.071 [0.038, 0.105] p=0.00003 |
| | 64 | | 0.049 [0.008, 0.089] p=0.01800 | 0.074 [0.039, 0.109] p=0.00003 | 0.082 [0.046, 0.118] p<1e-5 | 0.095 [0.058, 0.133] p<1e-5 |

(b)

| | Sample Size ELIXR-B | 64 | 512 | 4096 | 32768 | 68801 | 674533 |
|---|---|---|---|---|---|---|---|
| CXR-14 | Sample size | | | | | | |

|  | SupCon |  |  |  |  |  |  |
|---|---|---|---|---|---|---|---|
| Airspace opacity | 674533 | -0.047 [-0.058, -0.035] p<1e-5 | -0.041 [-0.054, -0.029] p<1e-5 | -0.033 [-0.043, -0.024] p<1e-5 | -0.009 [-0.015, -0.003] p=0.00445 | | |
| | 32768 | -0.028 [-0.039, -0.017] p<1e-5 | -0.023 [-0.036, -0.010] p=0.00039 | -0.015 [-0.022, -0.008] p=0.00003 | | | 0.027 [0.019, 0.036] p<1e-5 |
| | 4096 | **0.010 [-0.001, 0.022] p=0.08267** | **0.015 [0.002, 0.028] p=0.02128** | | 0.048 [0.036, 0.059] p<1e-5 | | 0.066 [0.053, 0.078] p<1e-5 |
| | 512 | **0.018 [0.013, 0.024] p<1e-5** | | 0.031 [0.019, 0.044] p<1e-5 | 0.056 [0.043, 0.069] p<1e-5 | | 0.074 [0.060, 0.087] p<1e-5 |
| | 64 | | 0.041 [0.027, 0.055] p<1e-5 | 0.049 [0.036, 0.062] p<1e-5 | 0.073 [0.059, 0.088] p<1e-5 | | 0.091 [0.076, 0.107] p<1e-5 |
| Fracture | 674533 | **-0.057 [-0.141, 0.027] p=0.18287** | **-0.051 [-0.138, 0.036] p=0.25343** | **-0.021 [-0.093, 0.052] p=0.57805** | **0.003 [-0.070, 0.076] p=0.92861** | | |
| | 32768 | **-0.059 [-0.134, 0.016] p=0.12386** | **-0.053 [-0.128, 0.023] p=0.17258** | **-0.022 [-0.104, 0.060] p=0.59567** | | | 0.014 [-0.056, 0.083] p=0.69461 |
| | 4096 | **-0.014 [-0.070, 0.043] p=0.63317** | **-0.007 [-0.100, 0.085] p=0.87417** | | 0.047 [-0.022, 0.116] p=0.18368 | | 0.059 [-0.016, 0.133] p=0.12069 |
| | 512 | **-0.006 [-0.087, 0.074] p=0.87682** | | 0.030 [-0.048, 0.109] p=0.45057 | 0.054 [-0.029, 0.137] p=0.20016 | | 0.066 [-0.020, 0.153] p=0.13367 |
| | 64 | | 0.028 [-0.064, 0.121] p=0.54877 | 0.058 [-0.046, 0.163] p=0.27171 | 0.082 [-0.008, 0.173] p=0.07281 | | 0.095 [-0.003, 0.193] p=0.05868 |
| Pneumothorax | 674533 | -0.232 [-0.272, -0.191] p<1e-5 | -0.114 [-0.149, -0.079] p<1e-5 | -0.104 [-0.133, -0.075] p<1e-5 | -0.039 [-0.062, -0.016] p=0.00107 | | |
| | 32768 | -0.172 [-0.209, -0.134] p<1e-5 | -0.054 [-0.093, -0.015] p=0.00676 | -0.045 [-0.075, -0.014] p=0.00428 | | | 0.066 [0.044, 0.088] p<1e-5 |
| | 4096 | -0.092 [-0.131, -0.053] p<1e-5 | **0.026 [-0.012, 0.064] p=0.18491** | | 0.101 [0.073, 0.130] p<1e-5 | | 0.146 [0.114, 0.178] p<1e-5 |
| | 512 | **0.010 [-0.038, 0.059] p=0.67914** | | 0.137 [0.097, 0.178] p<1e-5 | 0.203 [0.160, 0.246] p<1e-5 | | 0.248 [0.209, 0.287] p<1e-5 |
| | 64 | | 0.178 [0.126, 0.231] p<1e-5 | 0.188 [0.148, 0.227] p<1e-5 | 0.253 [0.204, 0.303] p<1e-5 | | 0.298 [0.256, 0.341] p<1e-5 |
| Consolidation | 68801 | -0.051 [-0.061, -0.041] p<1e-5 | -0.021 [-0.030, -0.012] p<1e-5 | -0.010 [-0.016, -0.004] p=0.00222 | **0.004 [-0.000, 0.008] p=0.06665** | | |
| Consolidation | 32768 | -0.050 [-0.060, -0.039] p<1e-5 | -0.019 [-0.027, -0.012] p<1e-5 | -0.009 [-0.014, -0.003] p=0.00274 | | 0.004 [-0.000, 0.009] p=0.08012 | |
| Consolidation | 4096 | -0.032 [-0.042, -0.022] p<1e-5 | **-0.002 [-0.009, 0.005] p=0.60796** | | 0.023 [0.016, 0.030] p<1e-5 | 0.022 [0.015, 0.028] p<1e-5 | |
| Consolidation | 512 | -0.014 [-0.019, -0.009] p<1e-5 | | 0.027 [0.018, 0.036] p<1e-5 | 0.041 [0.031, 0.050] p<1e-5 | 0.040 [0.030, 0.049] p<1e-5 | |
| Consolidation | 64 | | 0.070 [0.056, 0.083] p<1e-5 | 0.080 [0.069, 0.092] p<1e-5 | 0.094 [0.081, 0.107] p<1e-5 | 0.093 [0.080, 0.106] p<1e-5 | |
| Pleural effusion | 68801 | -0.090 [-0.097, -0.083] p<1e-5 | -0.073 [-0.079, -0.067] p<1e-5 | -0.033 [-0.038, -0.028] p<1e-5 | -0.004 [-0.006, -0.001] p=0.00524 | | |
| Pleural effusion | 32768 | -0.082 [-0.089, -0.075] p<1e-5 | -0.065 [-0.071, -0.059] p<1e-5 | -0.025 [-0.030, -0.021] p<1e-5 | | 0.011 [0.008, 0.014] p<1e-5 | |
| Pleural effusion | 4096 | -0.051 [-0.057, -0.045] p<1e-5 | -0.033 [-0.039, -0.028] p<1e-5 | | 0.036 [0.031, 0.041] p<1e-5 | 0.043 [0.038, 0.047] p<1e-5 | |
| Pleural effusion | 512 | **0.001 [-0.006, 0.007] p=0.82577** | | 0.058 [0.052, 0.064] p<1e-5 | 0.087 [0.080, 0.094] p<1e-5 | 0.094 [0.087, 0.101] p<1e-5 | |
| Pleural effusion | 64 | | 0.046 [0.041, 0.051] p<1e-5 | 0.086 [0.079, 0.093] p<1e-5 | 0.115 [0.108, 0.123] p<1e-5 | 0.122 [0.115, 0.129] p<1e-5 | |
| Pulmonary edema | 68801 | -0.142 [-0.157, -0.127] p<1e-5 | -0.090 [-0.101, -0.078] p<1e-5 | -0.021 [-0.028, -0.014] p<1e-5 | **-0.003 [-0.007, 0.002] p=0.24640** | | |
| Pulmonary edema | 32768 | -0.132 [-0.146, -0.117] p<1e-5 | -0.079 [-0.090, -0.068] p<1e-5 | -0.010 [-0.017, -0.003] p=0.00367 | | 0.013 [0.008, 0.017] p<1e-5 | |
| Pulmonary edema | 4096 | -0.100 [-0.117, | -0.048 | | 0.039 [0.030, | 0.044 [0.035, | |
| | | -0.084] p<1e-5 | [-0.062, -0.034] p<1e-5 | | 0.048] p<1e-5 | 0.053] p<1e-5 | |
| | 512 | -0.030 [-0.050, -0.010] p=0.00342 | | 0.091 [0.076, 0.106] p<1e-5 | 0.110 [0.095, 0.124] p<1e-5 | 0.114 [0.099, 0.129] p<1e-5 | |
| | 64 | | 0.053 [0.036, 0.069] p<1e-5 | 0.121 [0.107, 0.136] p<1e-5 | 0.140 [0.124, 0.155] p<1e-5 | 0.144 [0.129, 0.159] p<1e-5 | |

# Supplementary figures

**Supplementary figure 1: Suspect positive or negative ground truth MIMIC labels as identified by the query to Med-PaLM 2 truthed by a board certified radiologist.** Green represents correctly identified errors or lack of labels in MIMIC labels by Med-PaLM 2. Blue represents correct MIMIC labels. Red represents errors identified in both MIMIC and MedPalm2 based labels while gray represents indeterminate labels. A total of 1568 labels were flagged, of which 1092 were modified.

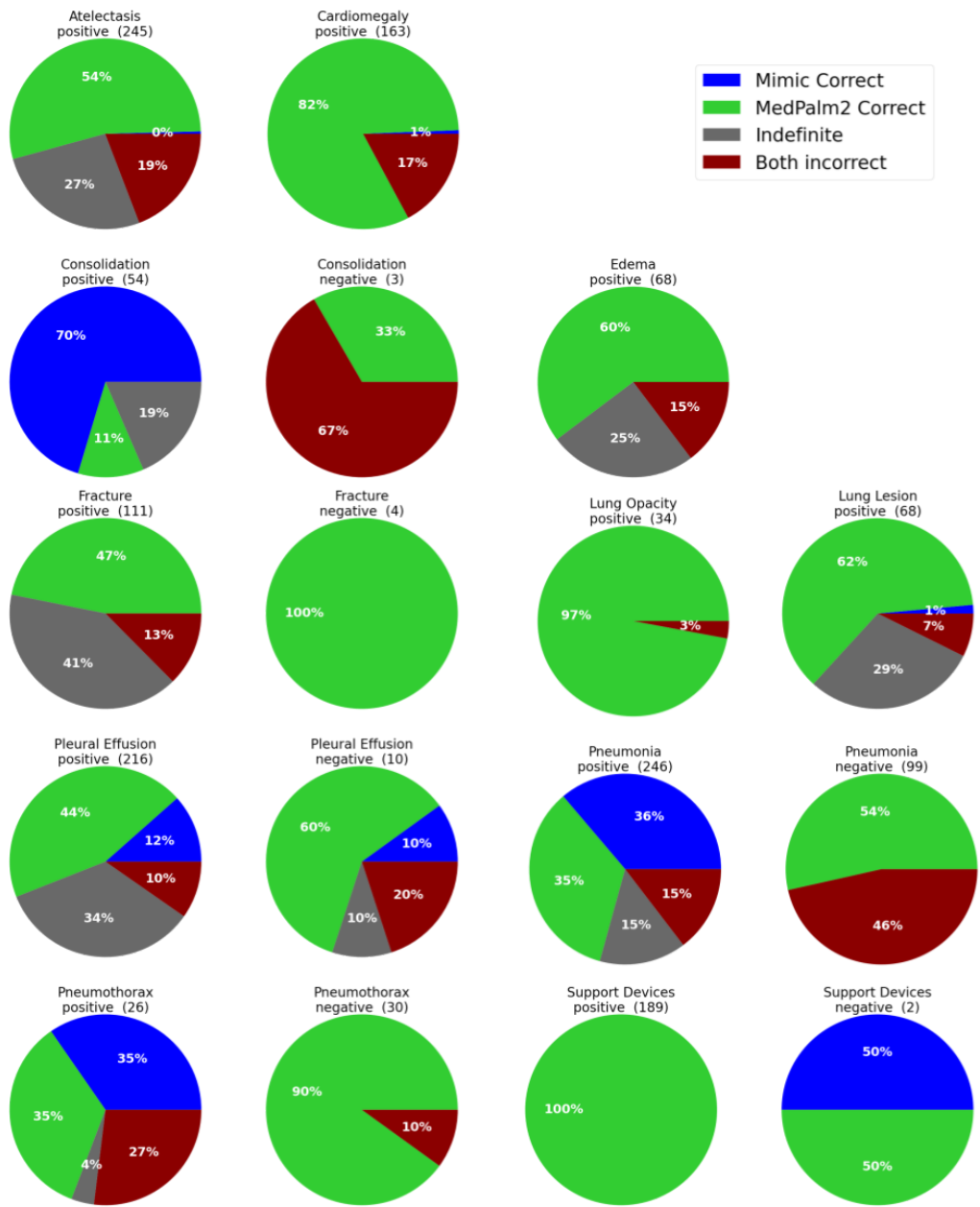

**Supplementary figure 2: Data-efficient classification performance between models per-dataset and per-finding using different training set sample size.** Evaluation on (a) CheXpert, (b) CXR-14. Significance test results are available in Supplementary tables 9, 10.

(a)

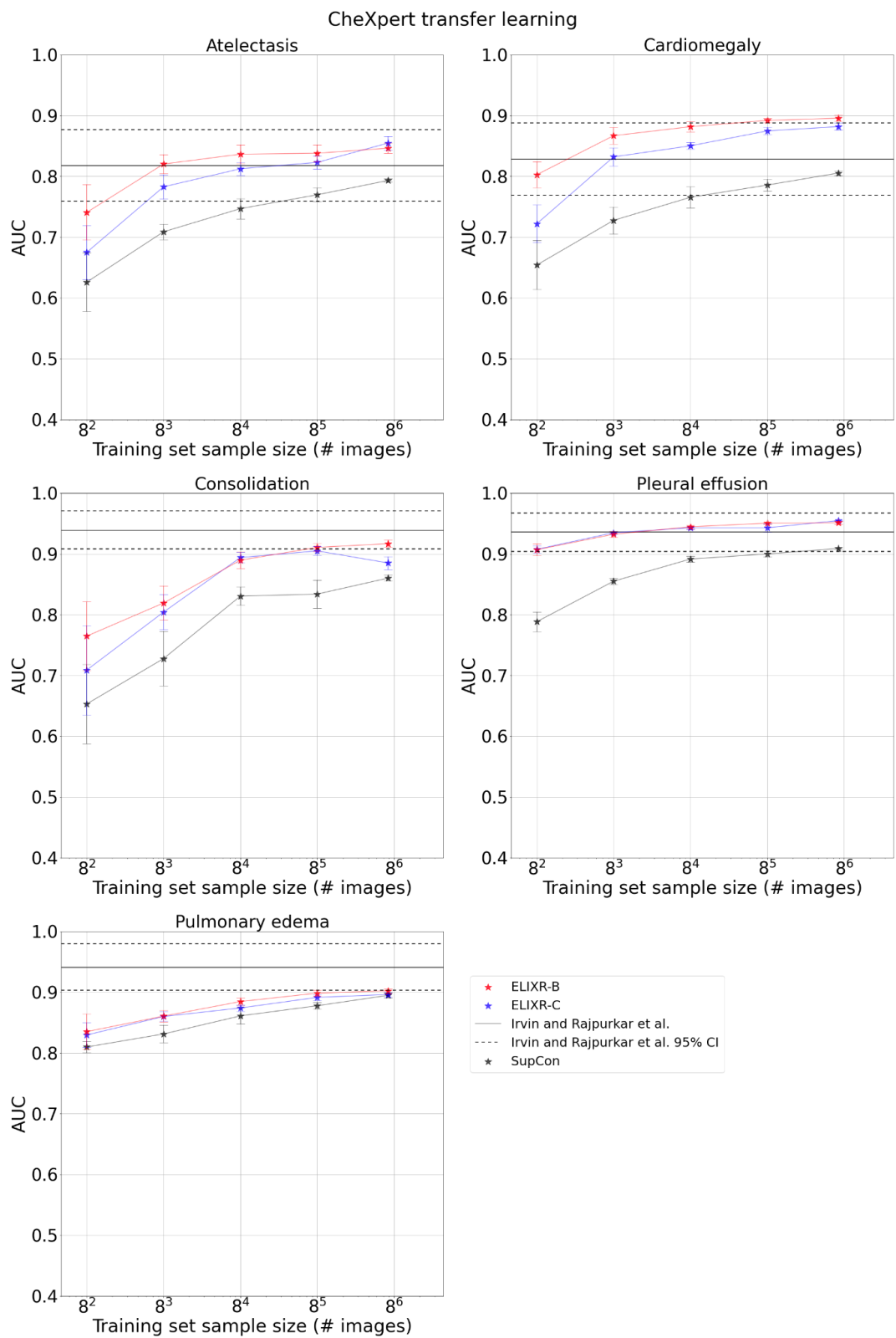

(b)

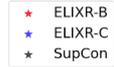
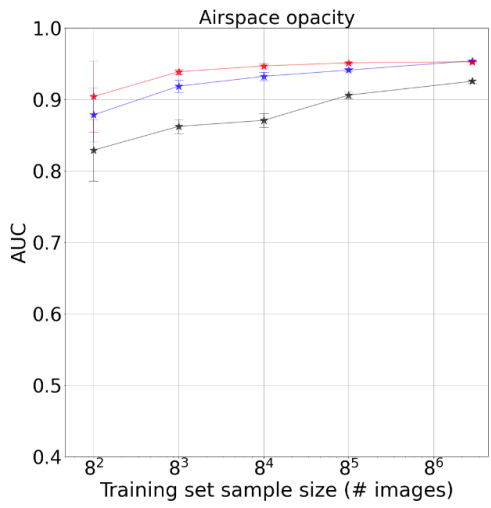
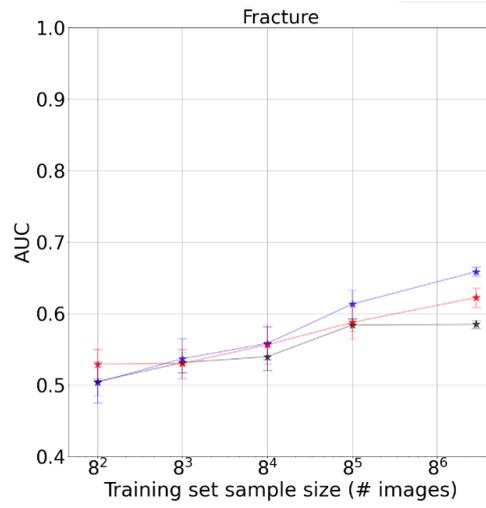
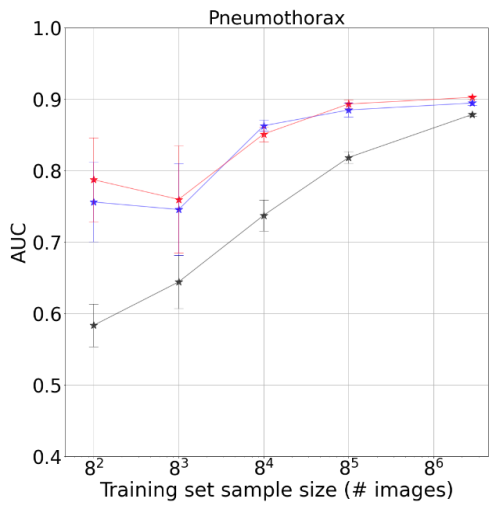
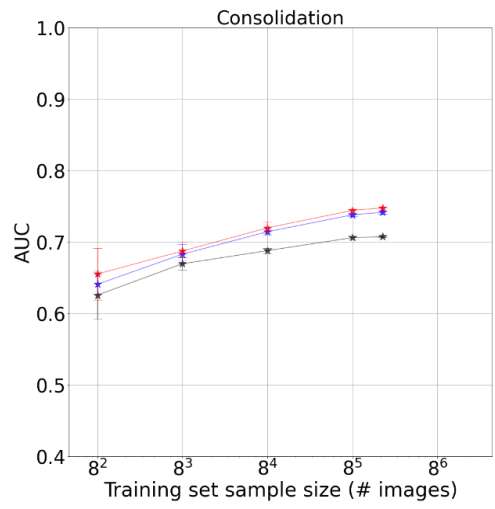
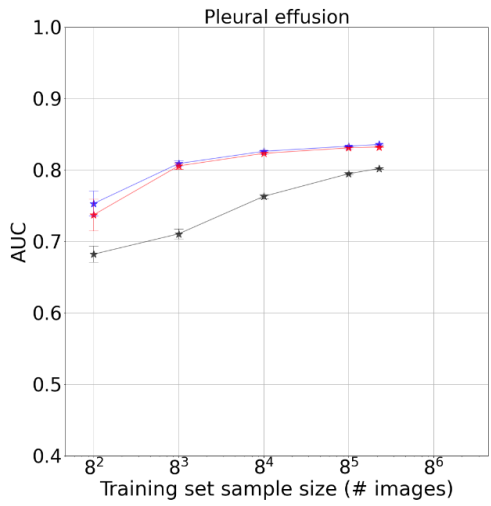
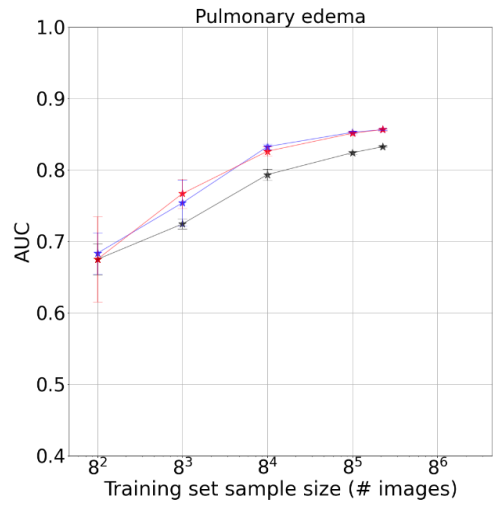